\documentclass[format=acmsmall, review=false, screen=true]{acmart}

\usepackage{booktabs} 
\usepackage[utf8]{inputenc}
\usepackage[ruled]{algorithm2e} 
\usepackage{multirow}
\usepackage[normalem]{ulem}
\useunder{\uline}{\ul}{}
\usepackage{tabularx}
\usepackage{longtable}
\usepackage{xcolor}
\usepackage{graphicx}
\definecolor{teal}{rgb}{0.13, 0.67, 0.8}

\SetAlFnt{\small}
\SetAlCapFnt{\small}
\SetAlCapNameFnt{\small}
\SetAlCapHSkip{0pt}
\IncMargin{-\parindent}
\usepackage[utf8]{inputenc}
\usepackage[english]{babel}
 \usepackage{booktabs}

\usepackage{makecell}
\usepackage{diagbox}
\usepackage{rotating}
\usepackage{boldline,multirow}
\usepackage{afterpage}
\usepackage{verbatim}

\begin{document}

\title[Deep learning advances on 3D data]{A survey on Deep Learning Advances on Different 3D Data Representations}

\author{Eman Ahmed}
\orcid{1234-5678-9012-3456}
\affiliation{%
  \institution{SnT, University of Luxembourg}
  \streetaddress{}
  \city{}
  \state{}
  \postcode{}
  \country{}}
\email{eman.ahmed@uni.lu}

\author{Alexandre Saint}
\affiliation{%
  \institution{SnT, University of Luxembourg}
  \city{}
  \country{}
}
\email{alexandre.saint@uni.lu}

\author{Abdelrahman Shabayek}
\affiliation{%
  \institution{SnT, University of Luxembourg}
  \city{}
  \country{}
}
\email{abdelrahman.shabayek@uni.lu}

\author{Kseniya Cherenkova}
\affiliation{%
  \institution{SnT, University of Luxembourg; Artec group, Luxembourg}
  \city{}
  \country{}
}
  \email{kcherenkova@artec-group.com}

\author{Rig Das}
\affiliation{%
  \institution{SnT, University of Luxembourg}
  \city{}
  \country{}
}
\email{rig.das@uni.lu}

\author{Gleb Gusev}
\affiliation{%
  \institution{Artec group}
  \city{}
  \country{Luxembourg}
}
  \email{gleb@artec-group.com}

\author{Djamila Aouada}
\affiliation{%
  \institution{SnT, University of Luxembourg}
  \city{}
  \country{}
}
\email{djamila.aouada@uni.lu}

\begin{abstract}
 $ \textbf{Abstract:} $  3D data is a valuable asset the computer vision filed as it provides rich information about the full geometry of sensed objects and scenes. Recently, with the availability of both large 3D datasets and computational power, it is today possible to consider applying deep learning to learn specific tasks on 3D data such as segmentation, recognition and correspondence. Depending on the considered 3D data representation, different challenges may be foreseen in using existent deep learning architectures.
In this work, we provide a comprehensive overview about various 3D data representations highlighting the difference between Euclidean and non-Euclidean ones. We also discuss how Deep Learning methods are applied on each representation, analyzing the challenges to overcome.

\end{abstract}

%
%

\begin {comment}
\end {comment}



%
%


\maketitle

\renewcommand{\shortauthors}{E. Ahmed et al.}

 The remarkable advances in the \emph{Deep Learning} (DL) architectures on 2D data has been coupled with notable successes in the computer vision field by achieving impressive results in many tasks such as: classification~\cite{krizhevsky2012imagenet}, segmentation~\cite{long2015fully,noh2015learning,saito2016real}, detection and localization~\cite {sermanet2013overfeat},  recognition~\cite{he2016deep} and scene understanding~\cite{farabet2013learning}. The key strength of deep learning architectures are in their ability to progressively learn discriminative hierarchical features of the input data. Most of the DL architectures are already established on 2D data~\cite{krizhevsky2012imagenet}. 
DL architectures on 2D data showed the requirement for large amount of training data. Due to this fact, applying DL on the 3D domain was not as effective as 2D.
Fortunately, with the latest advances in 3D sensing technologies and the increased availability of affordable 3D data acquisition devices such as structured-light 3D scanners~\cite{Geng11} and time-of-flight cameras~\cite{Hansard2012}, the amount of the available 3D data has tremendously increased. 3D data provides rich information about the full geometry of 3D objects. Driven by the breakthroughs achieved by DL and the availability of 3D data, the 3D computer vision community has been actively investigating the extension  of DL architectures to 3D data. 

3D data can have different representations where the structure and the geometric properties vary from one representation to another. 
In this paper, we study DL techniques employed on different 3D data representations in detail, classifying them into Euclidean and Non-Euclidean. 
3D Euclidean data has an underlying grid structure that allows for a global parametrization and a common system of coordinates. These properties make extending the already-existing 2D DL paradigms to 3D data a straightforward task, where the convolution operation is kept the same as 2D. 3D Euclidean data is more suitable for analyzing rigid objects where the deformations are minimal~\cite{sinha2016deep,brock2016generative,maturana2015voxnet} such as voxel data for simple objects~\cite{wu20153d}. On the other hand, 3D non-Euclidean data do not have the gridded array structure where there is no global parametrization. Therefore, extending classical DL techniques to such representations is a challenging task; however, understanding the structure of these representations is important in analyzing non-rigid objects for various applications such as segmentation tasks~\cite{maron2017convolutional} on human body models along with point-to-point correspondence~\cite{verma2018feastnet,fey2017splinecnn,monti2017geometric}. For the sake of expanding the scope of DL architectures and easing the applicability of DL models to 3D data, one needs to understand the structural properties of different representations of 3D data, where the focus of this paper is directed.

DL on 3D data has become a field in itself, with regular activities in top computer vision scientific venues. Different terminologies are used across different papers. In~\cite{bronstein2017geometric}, Bronstein et al. focus specifically on non-Euclidean data and refer to it as \textit{geometric data}. They initiate the term \textit{geometric deep learning} to refer to DL techniques applied on non-Euclidean data. The main purpose of this paper is to extensively study DL advances on all types of 3D data representations classifying them into Euclidean and non-Euclidean representations. 
In this paper, we provide a comprehensive overview about DL recent advances on different 3D data representations while emphasizing the challenges that are emerging from the structural differences between these data representations. We study some questions that drive the current research community in this area.
Mainly: (i) What is the relationship between the design of DL architecture and the 3D data representation? (ii) What is the optimal set of applications that can be achieved by each representation?\\
The present survey is different from a recent one given in~\cite{ioannidou2017deep}, where the focus was mostly on 3D DL applications in computer vision rather than different on the relationship between 3D data representations and DL. Moreover,~\cite{ioannidou2017deep} covers DL applications on Euclidean data while not considering DL models for non-Euclidean data. Hence, the current work is more comprehensive and covers the structural aspect of 3D data along with the employed DL models. We summarize the contribution of this paper as follows: 
\begin{itemize}
	\item A comprehensive survey about recent advances of DL on various 3D data representations, while distinguishing between the Euclidean and non-Euclidean representations. We also highlight the differences between different DL models applied on each category along with the challenge imposed to the structure of the data. 
	\item Analysis of the applications and tasks that can be achieved by each data representation are provided using a comparison between different DL techniques for the same task.
	\item Insights and analysis of the DL models that are employed and an introduction to new research directions. 
\end{itemize}
In order to keep this paper self-contained, in Section ~\ref{sec:data_Rep}, we start by over-viewing various 3D data representations and highlighting the structural differences between Euclidean and non-Euclidean representations. In Section~\ref{DLon_3D}, a study on recent advances of DL approaches over the two previously discussed data representations are provided by highlighting the main differences between the network models and the features that are learned by each DL paradigm. Section~\ref{analysis} provides an analysis on the evolution of the use of DL on 3D data and the motivations behind moving from one approach to another by highlighting the limitations of each model and possible areas of improvements. In Section~\ref{3D Computer Vision tasks}, we overview some of the most popular computer vision tasks specially 3D recognition~\ref{3D recognition} and 3D correspondence~\ref{coress_sec}. Finally, in Section~\ref{conclusion}, the major challenges in this field are discussed along with the concluding remarks.  


\section{Overview of 3D data representations} \label{sec:data_Rep}
\begin{figure*}[t]
	\centering
	\includegraphics[width=1.0\textwidth]{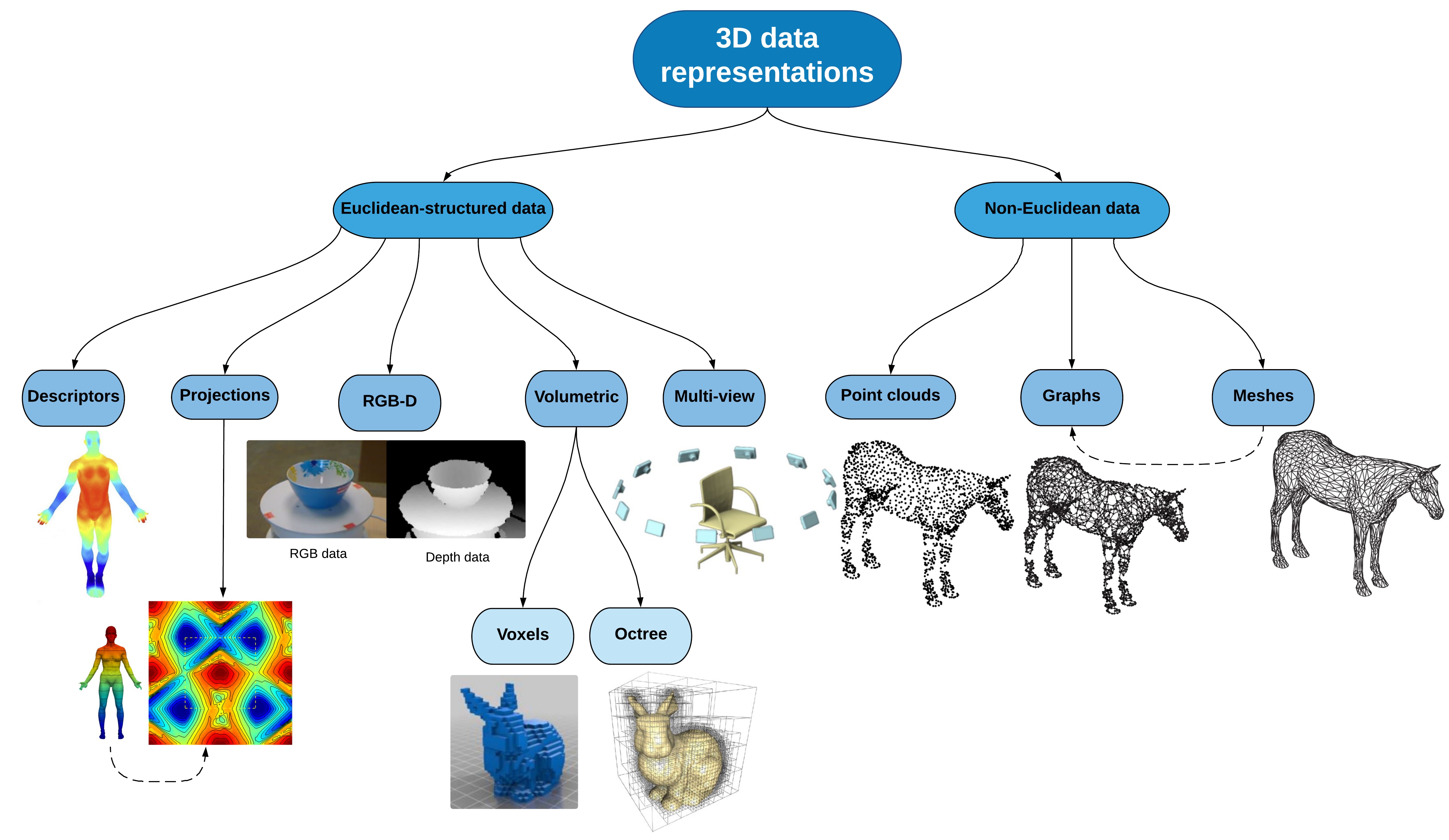}\caption{Various representations for 3D data: Euclidean representation (Descriptors~\cite{dense}, Projections~\cite{maron2017convolutional}, RGB-D~\cite{lai2011large}, Volumetric; voxels and octree~\cite{octree} and multi-view~\cite{su2015multi}) and Non-Euclidean representations (Point clouds, graphs and meshes)~\cite{discreteGeom}.}
	\label{fig:3D_data}
	\end {figure*}
	
	Raw 3D data captured by different scanning devices come in different forms that vary in both, the structure and the properties. In this section, we go through different representations of 3D data by categorizing them into two main families: Euclidean-structured data and non-Euclidean data. Previous efforts tried to address one or more 3D data representations, but not both categories~\cite {berger2013role,guo2014benchmark,chen2015utd,firman-cvprw-2016}. The present work intends to be more comprehensive and gives an in-depth information about different 3D data representations.


	\subsection{Euclidean data}
	Some 3D data representations have an underlying Euclidean-structure where the properties of the grid-structured data are preserved such as having a global parametrization and a common system of coordinates. The main 3D data representations that fall under this category are: descriptors, projections, RGB-D data, volumetric data and multi-view data. 
	
	\subsubsection{Descriptors}

	Generally speaking, shape descriptors are simplified representations of 3D objects to describe geometric or topological characteristics of the 3D shape. Shape descriptors can be obtained from the object's geometry, topology, surface, texture or any other characteristic or a combination of all~\cite{kazmi2013survey,zhang2007survey}. Shape descriptors can be seen as a signature of the 3D shape to ease processing and computations and allow for comparison among different 3D shapes. The nature and the meaning of this signature depend on the characteristic of the shape descriptor used and its definition. Kazmi et al.~\cite{kazmi2013survey} and Zhang et al.~\cite{zhang2007survey} have published comprehensive surveys about 3D shape descriptors. 
	
	Descriptors are usually combined with a learning-based model to extract more hierarchical discriminative features to better represent the shape. This will be further discussed in Section~\ref{Deep learning architectures on 3D data descriptors}.

	3D shape descriptors can be classified into two main categories; 
	(i) Based on whether the descriptor provides a local or global description of the shape. 
	Global descriptors provide a concise yet informative description for the whole 3D shape, e.g.~\cite{wohlkinger2011ensemble,aldoma2012our,marton2011combined} while local descriptors provide a more localized representation for smaller patches in the shape, e.g.~\cite{rusu2009fast,yi2016scalable,johnson1999using,guo2013rotational,bronstein2010scale}. \\(ii) Based on whether the nature of the geometric information captured is intrinsic or extrinsic. Intrinsic descriptors preserve the intrinsic geometric essence of the features described on the 3D shape such independently of any external references, as~\cite{aubry2011wave,bronstein2010scale,aouada2007ICIP,aouada2008CVPRW,aouada2010TIP,sun2009concise,bronstein2010scale}. Conversely, extrinsic descriptors describe the properties related to a specific realization of the 3D shape in the Euclidean space~\cite{chen2003visual,johnson1999using,ling2007shape,rusu2009fast,rusu2008aligning}.

	\subsubsection{3D data projections}
	
	Projecting 3D data into another 2D space is another representation for raw 3D data where the projection converts the 3D object into a 2D grid with specific features. The projected data encapsulates some of the key properties of the original 3D shape. The type of preserved features is dependent on the type of projection. Multiple projections have been proposed in the literature. Projecting 3D data into the spherical and cylindrical domains~\cite{cao20173d,shi2015deeppano} has been a common practice for representing the 3D data. Such projections, see Fig.~\ref{fig:3D_data}, represent 3D data in 2D while being invariant to rotations around the principal axis of the projection. This eases the processing of 3D data due to the Euclidean grid structure of the resulting projections and enables the usage of the well-researched learning models. 
	However, such representations are not optimal for complicated 3D computer vision tasks such as dense correspondence due to the information loss in projection~\cite{sinha2016deep}. 
	popular device            
	\subsubsection{RGB-D data}
	Representing 3D data as RGB-D images has become popular in the recent years, due to popular RGB-D sensors e.g., Microsoft's Kinect. As shown in Fig.~\ref{fig:3D_data}, RGB-D data provides a 2,5D information about the captured 3D object by providing the depth map (D) along with 2D color information (RGB). Besides being inexpensive, RGB-D data are simple yet effective representations for 3D objects to be used for different tasks such as identity recognition~\cite{erdogmus2013spoofing}, pose regression~\cite{fanelli2011real}, scene reconstruction~\cite{afzal2014rgb}, and correspondence~\cite{zollhofer2014real}. The number of available RGB-D datasets is huge compared to other 3D datasets such as point clouds or 3D meshes~\cite{firman-cvprw-2016}.

	\subsubsection{Volumetric data}
	3D data can be characterized as a regular grid in the three-dimensional space. Voxels are used to model 3D data by describing how the 3D object is distributed through the three-dimensions of the scene. Viewpoint information about the 3D shape can be encoded as well by classifying the occupied voxels into visible, occluded or self-occluded. Despite the simplicity of the voxel-based representation and its ability to encode information about the 3D shape, and its viewpoint, it suffers from some constraining limitations~\cite{xiang2015data}. Voxel-based representation is not always efficient because it represents both occupied and non-occupied parts of the scene, which establishes an enormous unnecessary need for memory storage. That is why voxel-based representation is not suitable for representing high-resolution data~\cite{abdul2007spatial,tatarchenko2017octree}.
	
	A more efficient 3D volumetric representation is octree-based~\cite {tatarchenko2017octree}, which is simply varying-sized voxels. Octree representation models 3D objects as a hierarchical data structure that models occupancy of the 3D object in the 3D scene~\cite{dong1996three} as shown in Fig.~\ref{fig:3D_data}. Octree representation is based on recursive decomposition of the root voxels similar to the quadtree structure~\cite {samet1984quadtree,abdul2007spatial}. The tree divides the 3D scene into cubes that are either outside or inside the object. Despite, the simplicity of forming 3D octrees, they are powerful in representing the fine details of 3D objects compared to voxels with less computations because of their ability to share the same value for large regions of space. However, both voxels and octree representations do not preserve the geometry of 3D objects in terms of the intrinsic properties of the shapes and the smoothness of the surface.

	\subsubsection{Multi-view data}
	3D data may be presented as a combination of multiple 2D images captured for the 3D object from different view points~\cite{zhao2017multi} as shown in Fig.~\ref{fig:3D_data}. Representing 3D data in this manner allows learning multiple feature sets for reducing the noise effect, incompleteness, occlusion and illumination problems on the captured data. Learning 3D data from the rendered 2D multi-view images of the same object aims to learn a function modelling each view separately and then jointly optimize all the functions to represent the whole 3D shape and to allow generalizing to other 3D shapes. However, the question of how many views are sufficient to model the 3D shape is still open. Representing the 3D object with an insufficiently small number of views might not capture the properties of the whole 3D shape and might cause an over-fitting problem. In addition, too many views cause an unneeded computational overhead. However, learning well-represented multi-view data proved better performance over learning 3D volumetric data~\cite{su2015multi}.  
	
	Both volumetric and multi-view data are more suitable for analyzing rigid data where the deformations are minimal. A good example for such objects is $ModelNet$ dataset~\cite{wu20153d} which is composed of CAD models for primitive objects such as, chairs, tables, desks, etc. Unlike, highly deformable non-rigid objects like the \textit{BU4D-FE} 3D facial expressions dataset~\cite{yin20063d}.

	\subsection{{Non-Euclidean data}} \label{Non-Euclidean data}
	The second type of 3D data representations is the non-Euclidean data. This type of data does not have a global parametrization or a common system of coordinates. Also, it lacks a vector space structure~\cite {bronstein2017geometric}, which makes extending 2D DL paradigms not a straightforward task.

	Considerable efforts were directed towards learning such data representation and applying DL techniques on it. The main type of non-Euclidean data is point clouds, 3D meshes and graphs. These structures have several properties in common which will be discussed throughout this section. It is important to note that both point clouds and meshes can be seen as both Euclidean and non-Euclidean data depending on the scales on which the processing is taking place, i.e., globally or locally.
	Despite this dual nature, we chose to list them as part of the non-Euclidean data because even if this data looks like Euclidean locally, in practice, they suffer from infinite curvature and self-intersections. Also, depending on the scale and location at which one looks, this data has different dimensions ~\cite{bronstein2017geometric}. Moreover, processing such data usually happens on a global scale to learn the whole 3D object's features which is convenient for complex tasks such as recognition and correspondence.

	\subsubsection{3D Point clouds}	A point cloud can be seen as a set of unstructured 3D points that approximate the geometry of 3D objects. Such realization makes it a non-Euclidean geometric data representation. However, point clouds can also be realized as a set of small Euclidean subsets that have a global parametrization and a common system of coordinates and invariant to transformations such as translation and rotation. That is why the definition of the point cloud's structure depends on whether one is considering the global or the local structure of the object. Since, most of the learning techniques strive for capturing the global features of the object to perform complex tasks such as recognition, correspondence, matching or retrieval; we classified point clouds as non-Euclidean data. 
	
	Despite the ease of capturing point clouds using any of the available technologies such as Kinect and structured light scanners, processing them is a challenging task due to some problems related to their lack of structure and the acquirement process from the environment. The data-structure related problems usually emerge due to the absence of connectivity information in point clouds, which leads to ambiguity about the surface information. Motivated by the use of point clouds in multiple computer vision tasks e.g., 3D reconstruction~\cite {park2011high}, object recognition~\cite{rangel2017object} and vehicle detection~\cite{yan2018supervised}, a lot of work has been done on processing point clouds for noise reduction such as the work done in ~\cite{han2017guided} with a purpose of filtering 3D point clouds from noise while preserving geometric features.
	
	\subsubsection{3D meshes and graphs} \label{dataRep_3D meshes and graphs.}
	3D meshes are one of the most popular representations for 3D shapes. A 3D mesh structure consists of a set of polygons called faces described in terms of a set of vertices that describe how the mesh coordinates exist in the 3D space. These vertices are associated with a connectivity list which describes how these vertices are connected to each other. The local geometry of the meshes can be characterized as a subset of the Euclidean space following the grid-structured data~\cite{bronstein2017geometric}. However, on a global aspect, meshes are non-Euclidean data where the known properties of the Euclidean space are not well defined such as shift-invariance, operations of the vector space and the global parametrization system.    
	
	Learning 3D meshes is a challenging task because of two main reasons: DL methods have not been readily extended to such irregular representations. Besides, such data usually suffer from noise, missing data and resolution problems such as~\cite{cosmo2016shrec}. That is why many works are directed towards 3D shape completition and inpainting such as~\cite{han2017high,wang2017shape}. 3D meshes can also be presented as graph-structured data where the nodes of the graph correspond to the vertices of the mesh and the edges represent the connectivity between these vertices. Graphs can be directed or undirected. As will be discussed in Section~\ref{DLon_3D}, many recent works have been done for exploiting such data to learn the properties of 3D objects. Analyzing the spectral properties of the graphs enabled researchers to use the graph Laplacian eigen-decomposition to define a convolution-like operation on graphs or meshes converted to graphs~\cite{fey2017splinecnn}. Such a start opened the door for promising innovations in processing geometric data.

	\section{Deep learning architectures on different 3D data representations} \label{DLon_3D}
	Deep Learning has remarkably contributed on the computer vision field achieving state-of-the-art results on several 2D computer vision tasks ~\cite{krizhevsky2012imagenet,long2015fully,noh2015learning,saito2016real,sermanet2013overfeat,he2016deep,farabet2013learning}, DL started gaining popularity in the 3D domain attempting to make use of the rich 3D data available while considering its challenging properties. However, extending DL models to 3D data is not straightforward due to the complex geometric nature of 3D objects and the large structural variations emerging from having different 3D representations.
	
\begin{figure*}[t]
		\centering
		\includegraphics[width=1.0\columnwidth]{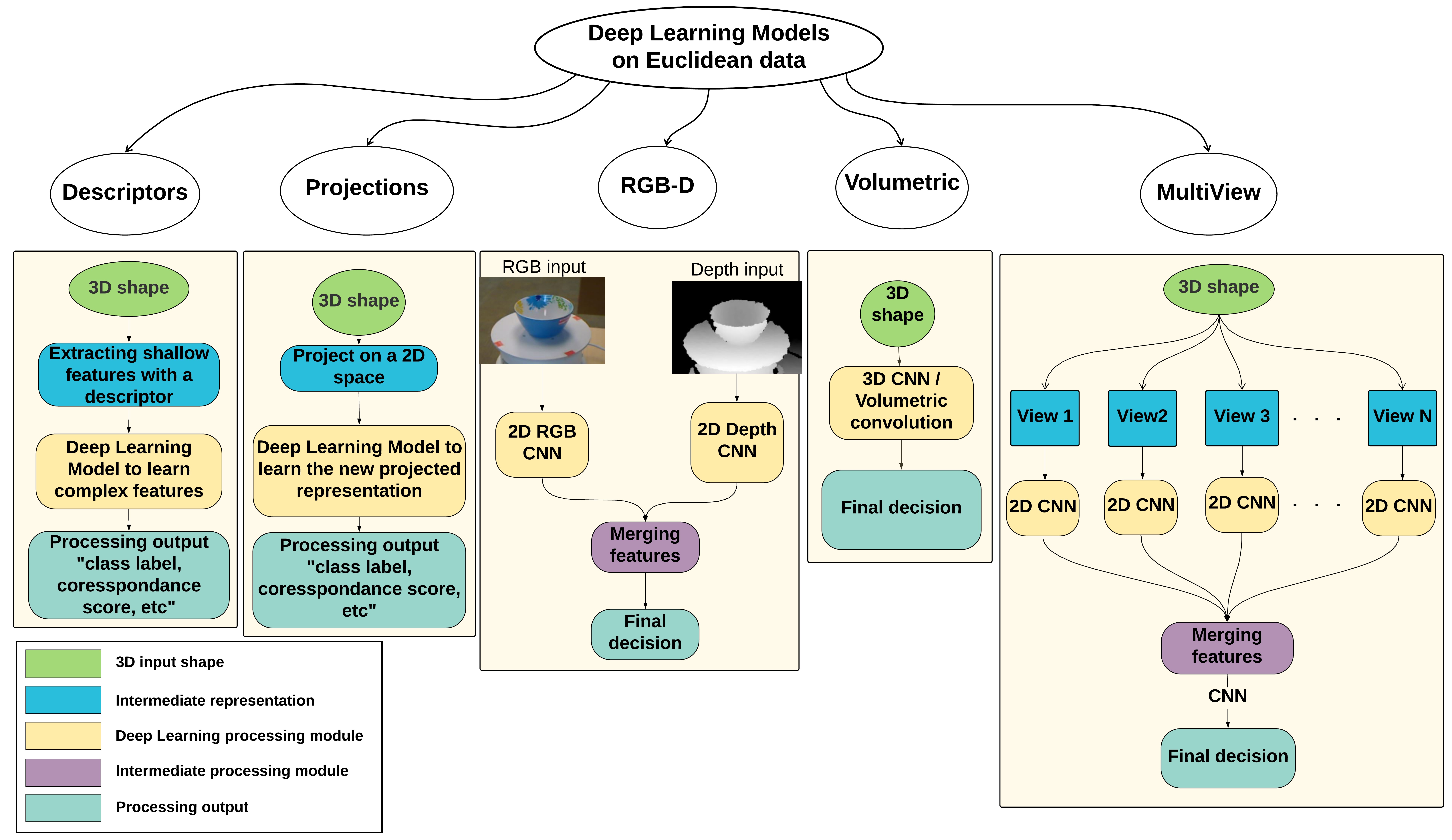}
		\caption{DL models on various Euclidean representations for 3D data.}
		\label{fig:taxonomy}
		\end {figure*}

	Having different 3D data representations has led researchers to pursue different DL routes to adapt the learning process to the data properties. In this section, we overview different DL paradigms applied on 3D data classifying them into two different families based on the data representation: DL architectures on Euclidean data and DL architectures on non-Euclidean data. An abstract illustration of the used methods is depicted in Fig.~\ref{fig:taxonomy}.

	\subsection{Deep learning architectures on 3D Euclidean-structured data} \label{Sec:DL_non}
	
	The first type of 3D DL approaches is the Euclidean approaches which operate on data with an underlying Euclidean-structure. Due to the grid-like nature of such data, the already established 2D DL approaches can be directly applied. Representing 3D data in a 2D way implies that the initial 3D representation is subjected to some processing to produce a simpler 2D representation on which the classical 2D DL techniques can operate. This processing evolved over years resulting in different 2D representations for 3D data with different properties and characteristics. DL architectures are adapted to each representation trying to capture the geometric properties of this data representation. 
	
	Initially, researchers made use of already established developments~\cite{tangelder2004survey,liu2013survey} to extract features from 3D data which produces grid-like shallow features where 2D DL models can be directly applied. However, considering the complexity of 3D data, this type of methods does not discriminatively learn the intrinsic geometric properties of the shape and might result in omitting significantly important information about the 3D object. This pushed towards exploiting the depth modality directly from the RGB-D data with DL models, which was effective in some applications, however, this type of data is not suitable for analyzing complex situations~\cite{firman-cvprw-2016}. This motivated researchers to apply DL approaches on 3D data directly such as volumetric data and multi-view data. In this section, we cover different DL architectures based on the aforementioned data representations discussing the main strengths and weaknesses of each model and the type of features learnt in each case.      
	
		\subsubsection{Deep learning architectures on 3D data descriptors} \label{Deep learning architectures on 3D data descriptors}
		
		Low-level descriptors have been used as a significant part of the learning process for 3D data. Although multiple handcrafted low-level descriptors were proposed in the literature such as~\cite{tangelder2004survey} and~\cite{liu2013survey}, but they suffer from several significant limitations, as they cannot learn the discriminative features from 3D shapes. Hence, the global and local structure of the 3D shapes cannot be preserved. Fortunately, DL models are effective in learning hierarchical discriminative features that can generalize well to other unseen data. That is why low-level descriptors have been combined with DL architectures to learn more informative high-level features of the 3D object as shown in the descriptors processing pipeline in Fig.~\ref{fig:taxonomy}. This common practice was followed by Liu et al.~\cite{liu2014high} for the purpose of learning high-level features in order to be used for classification and retrieval tasks. Visual ``Bag-of-Words (BoWs)" encode low-level feature representations, which are then fed into the ``Deep Belief Networks (DBNs)" for learning the high-level semantic features of the input. The results of the 3D retrieval and classification experiments showed that the learnt features are discriminative against inter-class variations achieving better results than the classical BoW low-level features. In~\cite{Bu2014LearningHF}, Bu et al. proposed a three-stage pipeline to learn the geometric properties of 3D shapes. Their main idea was to use low-level features extracted to build middle-level position-independent geometric features on which a DL model can be employed to learn the hierarchical high-level features of the 3D shape. In this method, both ``Scale-Invariant Heat Kernel Signature (SI-HKS)"~\cite{bronstein2010scale} and ``Average-Geodesic Distance (AGD)" were employed as to learn low-level features. Then, the Spatially Sensitive ``Bag-of-Features (SS-BoF)" was employed to learn spatially close words and the relationship between them. Finally, DBNs were used on the SS-BoWs to learn high-level features. Experiments on 3D retrieval and recognition demonstrated significant improvements compared to using low-level descriptors only. In~\cite{bu2015local}, Bu et al. extended their work in~\cite{Bu2014LearningHF} to a GPU based implementation to accelerate the computations and used it for the tasks of correspondence and symmetry detection where the proposed framework proved to achieve better performance. 
		
		Motivated by the HKS performance in extracting low-level features, Xie et al.~\cite{xie2015projective} used HKS to be employed as a low-level descriptor at multi-scales. The result was fed to AEs to learn discriminative features for 3D retrieval. As an enhancement for the features representations, "\textit{Fisher Discriminative Analysis} (FDA)" was also applied. Experiments proved the robustness of this model against deformations. In~\cite{han2017mesh}, Han et al. proposed ``\textit{Mesh Convolutional Restricted Boltzmann Machines} (MCRBMs)" to learn the hierarchical discriminative features of 3D meshes. The proposed model was able to learn both the global and local features of 3D objects. The structure of the local features was preserved by using the "\textit{Local Function Energy Distribution} (LFED)". An extension, a deeper model composed of multiple stacked MCRBMs, was tested in the context of shape retrieval and correspondence. This model has outperformed the current state-of-the-art techniques like~\cite{wu20153d} and~\cite{bronstein2010scale}. 
		
		Most of the DL models employed in the previous methods fall under the unsupervised learning methods category because supervised methods tend to learn hierarchical abstractions about the raw data. However, presenting 3D data with descriptors is indeed a form of abstraction. That is why supervised methods might not produce informative features because it would learn abstractions of the abstractions which might lead to a loss of the actual properties of the shapes if the descriptor representation is very simple/abstract. That is where unsupervised methods are more suitable for such representation to learn the hidden patterns or grouping in the input data. However, in some cases, the descriptors can provide rich information on which the convolution operation can be effective to learn the hierarchical features of the input representations such as~\cite{han2017mesh}. These methods can still be combined with unsupervised models. In short, the choice of the DL model on descriptor representations is dependent on how rich the descriptor is.                                                                              
		
		\subsubsection{Deep learning architectures based on 3D data projections}  \label{Deep learning architectures based on 3D data projections}
		One of the first attempts towards learning the features of 3D data by projecting them into 2D planes was presented by Zhu et al. in~\cite{zhu2016deep}. The proposed pipeline started by some data pre-processing where translation, scaling and pose normalization were applied on each 3D model. Then, various 2D projections were applied on each processed 3D model to feed it to a stack of RBMs to extract the features of different projections. In order to learn a global representation of 3D objects to be used for the retrieval task, An AE was employed. Experiments showed that this framework performed better than global descriptor-based techniques. The performance was boosted by combining local representations with the learned global ones. In this context, Shi et al.~\cite{shi2015deeppano} proposed \textit{DeepPano}. DeepPano refers to extracting 2D panoramic views from 3D objects by employing a cylindrical projection around the principal axis of the 3D object. 2D classical CNN architecture was used to train the model. In order to achieve rotation invariance around the principal axis, row-wise max-pooling layer was used between the ``\textit{Convolution} (Conv)" and ``\textit{Fully Connected} (FC)" layer. The proposed network consisted of four Conv layers, one row-wise max-pooling layer, two FC layers and one softmax layer inserted at the very end of the network. Proposed model was tested on 3D object recognition and retrieval tasks where it proved its effectiveness in comparison with previous models. Sinha et al. in~\cite{sinha2016deep} proposed \textit{geometry images} where 3D objects were projected into 2D grid so that the classical 2D CNNs can be employed. The proposed method created a planner parametrization for 3D objects by using authalic (area conserving) parametrization on a spherical domain to learn the 3D shapes surfaces. Then, the constructed geometry images were served as inputs to classical CNN architecture to learn the geometric features of the 3D objects. As a pre-processing step, data augmentation, scaling, rotation and translation operations were performed for increasing the size of the training data and for providing some variety. Geometry images were tested on various datasets: \textit{ModelNet10}~\cite{wu20153d}, \textit{ModelNet40}~\cite{wu20153d}, \textit{McGill11}, \textit{McGill2}, \textit{SHREC1} and \textit{SHREC2} on classification and retrieval tasks. Results showed that geometry images can produce comparable results in comparison with the state-of-the-art methods~\cite{wu20153d,su2015multi,shi2015deeppano}.
		
		\par Motivated by the results achieved by the projection methods, Cao et al.~\cite{cao20173d} proposed using a spherical domain projection to project 3D objects around their barycenter producing a set of cylindrical patches. The proposed model uses the projected cylindrical patches as an input to pre-trained CNNs. Two complementary projections were also used to better capture the 3D features. The first complementary projection captures the depth variations of the 3D shapes while the second learns the contour information embedded in different projections from different angles. The proposed model was used for 3D object classification task where it was tested on multiple datasets producing comparable results to previous methods. Similar to the previous work, Sfikas at al.~\cite{sfikas2017exploiting} proposed to represent 3D objects as panoramic views extracted from normalized posed 3D objects. Initially, 3D objects are preprocessed to normalize their poses using the framework of ``\textit{Pose Normalization of 3D Models via Reflective  Symmetry on Panoramic Views (SymPan)"}~\cite{sfikas2017ensemble} method. Then panoramic views are extracted to be combined and fed to CNN to perform classification and retrieval tasks. While being similar to the previous models, this method enhanced the accuracy when tested on \textit{ModelNet10} and \textit{ModelNet40} datasets. An extension to this model was introduced in~\cite{sfikas2017ensemble} where an ensemble of CNNs was used for the learning process. This extension produced very high results when tested on the aforementioned datasets.

		Projection-based representations are simple yet effective for learning 3D objects using 3D DL methods. The geometric properties of the shape is lost due to the projections; that is why in the previous work, researchers are trying to combine more than one projection representation to compensate for the missing information. Although 2D DL models can be directly applied on this representation, the networks usually require more fine-tuning than operating directly on the raw 3D data representation which might be tailored to the training data specifically and cause a shortage when tested on new unseen data.

		\subsubsection{Deep learning architectures on RGB-D data} \label{Deep learning architectures on RGB-D data}
		Due to the availability of RGB-D sensor data, multiple research efforts were directed towards leveraging the available data to exploit it for several tasks. One of the first approaches in this direction was proposed by Socher et al. in~\cite{socher2012convolutional}, where the authors presented a pipeline of convolution and recursive neural networks to process both the color images and depth channels of the RGB-D data. Two single-layer CNNs were employed in order to learn the representations of feature of the RGB-D input. The resulted descriptor was fed to multiple RNNs with random weights. Then, the results from the RNNs were combined and merged to be used as an input to a softmax classifier. This framework was used for household object classification where it showed accurate performance. Researchers continued to use the power of CNNs to learn the RGB-D data features. In~\cite{couprie2013indoor}, Couprie et al. proposed a multi-scale CNN indoor RGB-D scenes semantic segmentation. The proposed network learned the RGB-D data at multiple scales (three different scales). The network is mainly composed of two parallel CNNs where the first CNN is responsible for classifying the objects in the scene and the results of the second CNN are forwarded to a classifier to compute the class label prediction score. The final class label was decided by using the results from the classifier along with the segmented super-pixels of the scene. As a pre-processing step in this model, the channels of the RGB-D images are normalized to zero mean and the depth information is added as the fourth pixel to the RGB images and inputted to the CNNs. This method produced better accuracy by 6\% than previous models and computationally is very efficient. The takeaway from this method is that despite the simplicity of the use of the CNN combined with depth, the result is much better compared to handcrafted features. Also, this method demonstrates the substantial importance of the depth information for the segmentation application, as it is a key factor for separation among objects. However, the CNN here seems to learn the class object only without learning the geometry of the shape. 
		
		\begin{figure}[h]
			\centering
			\includegraphics[width=0.6 \columnwidth]{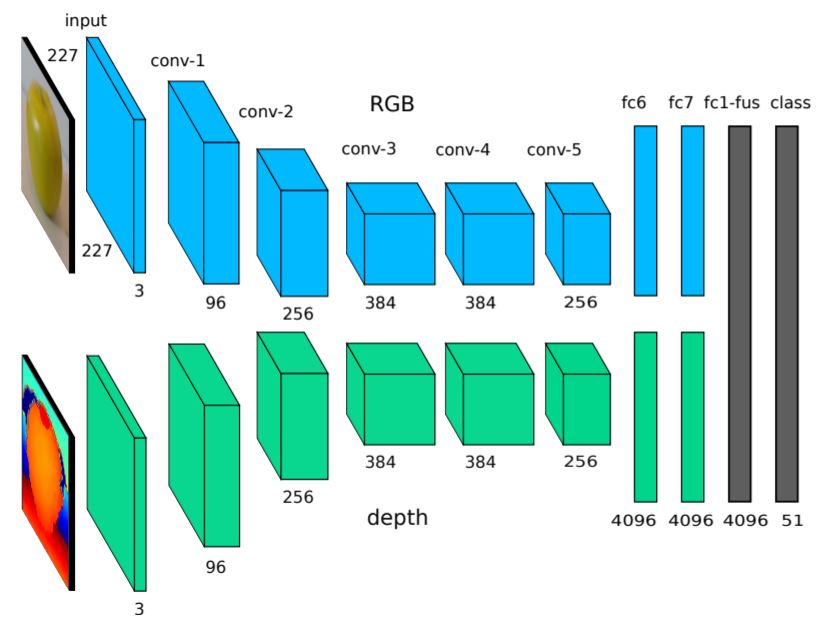}\caption{Two-stream CNNs on RGB-D data for the 3D object recognition task~\cite{eitel2015multimodal}. The figure reused from~\cite{eitel2015multimodal} with permission of authors.}
			\label{fig:RGBD}
			\end {figure}

			Inspired by the results of the two-stream network proposed in~\cite{couprie2013indoor}, researchers started exploiting the same concept introducing some novel modifications. Instead of using the networks for two different learning tasks (classification and segmentation),  researchers started processing the depth information and the color information separately using a different network, which started to be a common practice in processing RGB-D data as illustrated in the pipeline depicted in Fig.~\ref{fig:taxonomy} under the RGB-D data representation. Eitel et al. in~\cite{eitel2015multimodal} proposed to use a two-stream CNN on RGB-D data for 3D object recognition as shown in Fig.~\ref{fig:RGBD}. One CNN stream processes the RGB color information and the other stream is for processing the depth. Each of the two CNN has five Conv layers and two FC layers. Each network was trained separately and then the results were fused in the FC and softmax layers to decide on the object's class. This method outperformed previous existing methods and demonstrated a promising performance for object recognition in real-world noisy environments. Feng et al.~\cite{feng20163d} proposed an ensemble of AEs using only one RGB-D model for 3D object retrieval. Each AE was trained using the ``\textit{Stochastic Gradient Descent} (SGD)" algorithm on a dataset of different CAD models. Because of the difference between training data and the test data, the output scores of the AEs were then forwarded to what they called ``\textit{Domain Adaption Layer} (DAL)" to rank the retrieved scores. This method enhanced the performance in comparison with other related methods. 
			
			Alexandre~\cite{alexandre20163d} combined the concept of transfer learning and CNNs together to train four CNNs independently. In this model, each channel of the four channels in the RGB-D data was processed using a separate CNN and the weights were transferred from each network to another. Experiments indicated that this method boosted the performance which implies that the depth information carries valuable information about the 3D shape, which pushed Schwarz et al. in~\cite{schwarz2015rgb} to explore the transfer learning concept on RGB-D data for the object classification task. In this model, RGB-D data was rendered from a canonical perspective and the resulted depth was colored based on the distance from the object's centre. The CNN employed in this model was a pre-trained CNN for object categorization. The output of the network's two last layers was used as the object descriptor, which was forwarded to SVMs to learn the object's class. The pre-training stage enables the model to extract better features which helps boost the performance. 
			
			Deep learning proved to be effective in learning RGB-D data despite the simplicity of the models. Moreover, processing the depth channel separately conveys that the depth has valuable information about the 3D signal that contributes the whole learning process. However, these methods do not learn the full geometry of the 3D object and can only infer some of the 3D properties based on the depth. Later works we overview, examine the full volumetric representation of the 3D shape rather than using the flat 2D images of color and the depth information.

			\subsubsection{Deep learning architectures on volumetric data} \label{Deep learning architectures on volumetric data}
			Some efforts were directed towards processing the 3D volumetric representations of 3D objects to exploit the full geometry of the object. $ShapeNet$~\cite{wu20153d} is the first DL model exploiting the geometry of 3D objects represented as voxels. The input object is a 30x30x30 binary tensor indicating whether the voxel is part of the 3D object or not. A ``Convolutional Deep Belief Net (CDBN)"~\cite{lee2009convolutional} concept was adapted from 2D DL to model 3D objects. CDBNs can be seen as a variation of DBNs where convolution is also employed to benefit from the weight-sharing property to to reduce the number of parameters. That is why CDBNs are used to model and learn the joint probability distribution over voxels representing different object categories with a small number of parameters. $ShapeNet$ consists of five layers (the input layer, three convolution layers and the output layer). The proposed network was initially pre-trained. During the stage of pre-training, the network was trained in a layer-wise fashion where the first four layers are trained using the ``\textit{Contrastive Divergence}" method and the last layer was trained using the ``\textit{Fast Persistent Contrastive Divergence}". During the test stage, the input is provided as a single depth map for the 3D object, which is then converted into a voxel grid representation. $ShapeNet$ was tested for three different tasks: 3D shape classification, view-based recognition and next-best view prediction. Although ShapeNet was the first network to exploit 3D volumetric data directly in deep learning, it is imposing many constraints. The additional dimension in the convolution kernel results in a computationally intractable model that can hardly process large sized or high-resolution data. Also, the network is trained on isolated view of fixed size voxels without any additional information or background clutter which makes the learning process hard. Despite these limitations, this network produces impressive results 
			given that it is operating on low-resolution voxels. Also, besides presenting \textit{ShapeNet} in~\cite{wu20153d}, the authors also presented \textit{ModelNet} dataset which we describe in details in Section~\ref{3D datasets}. The availability of 3D CAD labelled models opened the door for more experiments and boosted the research in this area. 
			
			Maturana and Scherer in~\cite{maturana2015voxnet} exploited the concept of 3D convolution and proposed $VoxNet$ to perform 3D object recognition on different 3D data representations: RGB-D data, LIDAR point clouds and 3D CAD models. The convolution in $VoxNet$ followed the 2D Convolution except for the filter where a 3D filter was used instead of a 2D filter. The network architecture is composed of (the input layer, two Conv layers, a pooling layer and two FC layers). The input data was constructed as a volumetric occupancy grid of 32x32x32 voxels and it was fed to the network that was trained using SGD with momentum. Experiments showed that $VoxNet$ outperforms $ShapeNet$ when tested on $ModelNet10$, $ModelNet40$ and $NYU v2$ dataset for the classification task when the networks are trained from scratch. However, $ShapeNet$ outperforms $VoxNet$ when tested on $NYU v2$ when the pre-trained model for $ModelNet10$ is used. Motivated by the promising performance of $VoxNet$, Seaghat et al.~\cite{sedaghat2016orientation} modified the architecture of the $VoxNet$ to incorporate the orientation of the 3D object in the learning process. This enhanced the classification results on the $ModelNet10$ dataset. To learn the 3D data representations using unsupervised techniques, Sharma et al.~\cite{sharma16eccvw} proposed to use a ``\textit{Convolutional Volumetric Auto-Encoder} (VConv-DAE)" to learn the embedding of 3D objects in an un-supervised manner. VConv-DAE learns the volumetric representation from noisy data by estimating the occupancy grid of voxel data. In~\cite{wu2016learning}, Wu et al. presented the 3D-GAN to implicitly learn the 3D objects features the probabilistic latent space using the adversarial discriminator. The adversarial discriminator learns to capture the structure of 3D objects with the ability to identify whether it is real or synthesized. This carried a discriminative semantic information about 3D objects which is effective for modelling 3D objects and generating synthetic data as shown in the experiments. 
			
			The great advancements in the 2D very deep architectures motivated Brock et al.~\cite{brock2016generative} to adopt such models for 3D object classification on $ModelNet10$ and ModelNet40 datasets. The authors proposed \textit{Voxception-ResNet} (VRN) very deep model. As the name implies, VRN relied on Inception architectures~\cite{he2016deep}~\cite{szegedy2017inception}. In addition to adopting the batch normalization methods~\cite{ioffe2015batch,he2016deep} and stochastic network depth techniques~\cite{huang2016deep}, VRN is composed of 45 layers deep, which required data augmentation for the training process. VRN is similar to $VoxNet$ in a sense that they both adopt ConvNet with 3D filters but VRN is very deep compared to $VoxNet$, which achieved a significant improvement by 51.5\% in the classification task on $ModelNet$ datasets which marks the state-or-art performance on this dataset. Despite the remarkable performance of this method, it has a complex architecture and requires a significant amount of data augmentation to avoid the over-fitting problem that can result from the deep architecture of a small dataset. These constraints are limiting and can not easily be achieved. Xu and Todorovic proposed the beam search model to learn the optimal 3D CNN architecture to perform classification on $ModelNet40$ dataset~\cite{xu2016beam}. The proposed model identifies the 3D CNN number of nodes, number of layers, connectivity and the training parameters as well. The model starts with a fairly simple network (two Conv layers and one FC layer). The beam search method starts with this architecture and extends it to build the optimal 3D CNN model by either adding a new Conv filter or adding a new Conv layer. The beam search model is trained in a layer-wise fashion where the standard Contrastive Divergence method is used to train Conv layers are the  Fast Persistent Contrastive Divergence is used to train the FC layer. Once one layer is trained, the weights are then fixed and the activation parameters are transferred to the following layer. The proposed method produced significant results in the classification task on $ModelNet40$ dataset. 
			
			In an attempt to learn the 3D features at different scales using 3D CNNs, Song and Xiao in~\cite{DeepSlidingShapes} presented ``\textit{Deep Sliding Shapes}" model to perform 3D object recognition and classification on $ModelNet$ dataset. The authors converted depth maps from RGB-D scenes into 3D voxels using a directional ``\textit{Truncated Signed Distance Function} (TSDF)". Also, the 3D ``\textit{Region Proposed Network} (RPN)" was proposed to process the 3D object at two different scales and generate two 3D bounding boxes around the 3D object. This handles 3D data of different scales and sizes. The scene is pre-processed to obtain information about the orientation of the object to prevent ambiguity in the bounding boxes orientations. The power of this model comes from the TSDF representation which gives an informative representation about the 3D object's geoemetry instead of using the raw depth map. Also, the RGB values can be appended to the TSDF resulting in a compact representation. This model produced comparable results on $NYU v2$ dataset for the object detection tasks on various object classes
			
			Despite the effectiveness of 3D volumetric models, most of the current architectures requires a huge computational power due to the convolution operation and the large number of parameters. This is what motivated Zhi et al.~\cite{zhi2017toward} to propose \textit{LightNet}. \textit{LightNet} is a real-time volumetric CNN designed for the 3D object recognition task. The key power of \textit{LightNet} comes in two-fold. \textit{LightNet} leverages the power of multi-tasking to learn multiple features at the same time. Also, to achieve a faster convergence with less parameters, the batch normalization operation is utilized between the Conv operation and the activation. \textit{LightNet} includes two main learning tasks: the first one is for learning the class labels for each 3D voxel and the second task is to learn the orientation. \textit{LightNet} was tested on $ModelNet$ datasets for the classification task where it outperformed $VoxNet$ by approximately 24.25\% on $ModelNet10$ and 24.25\% on $ModelNet40$ with parameters parameters less than 67\% of $VoxNet$. This proves the strength and the capabilities of this proposed model. Further work tried to investigate more about the multi-view representation to incorporate all the geometric information about the scene from multi 2D views while using 2D DL models for processing which is computationally more plausible.

			\subsubsection{Deep learning architectures on multi-view data} \label{Deep learning architectures on multi-view data}
			
			Despite the effectiveness of volumetric deep learning methods, most of these approaches are computationally expensive because of the volumetric nature of the convolutional filters to extract the features which increase the computational complexity cubically with respect to the voxels resolution which limits the usage of 3D volumetric DL models. That is why exploiting multi-views of 3D objects is practical.  Indeed, it enables exploiting the already established 2D DL paradigms without the need to design a tailored model for 3D volumetric data with high computational complexity as shown in pipeline illustrated in Fig.~\ref{fig:taxonomy}. One of the first attempts to exploit 2D DL models for learning multi-view 3D data was presented by Leng et at. in~\cite{leng20143d} where DBN was employed on various view-based depth images to extract high-level features of the 3D object. A later-wise training manner was used to train the DBN using the ``\textit{Contractive Divergence}" method. The proposed model produced better results than the composite descriptors approach employed in~\cite{daras20103d}. Xie et al.~\cite{xie2015projective} proposed ``\textit{Multi-View Deep Extreme Learning Machine} (MVD-ELM)". The proposed MVD-ELM was employed on 20 multi-view depth images that were uniformly captured with a sphere at the centre of the 3D object. The proposed MVD-ELM contained Conv layers that had shared weights across all the views. The output activation weights were optimized according to the feature maps extracted. This work has been extended to be Fully Convolutional, resulting in (FC-MVD-ELM). FC-MVD-ELM was trained using the multi-view depth images to be tested for 3D segmentation. The predicted labels from the training stage were then projected back to the 3D object where the final result was smoothed using the graph cut optimization method. Both MVD-ELM and FC-MVD-ELM were tested on 3D shape classification and segmentation tasks and outperformed the previous work~\cite{wu20153d} and reduced the processing time significantly.  
			
			\par More research investigations were carried by Leng et al. to employ DL paradigms on multi-view 3D data. Leng et al. in~\cite{leng20153da} proposed an extension of classical AEs in a similar way to the CNN architecture. Their proposed framework is called ``\textit{Stacked Local Convolutional AutoEncoders} (SLCAE)". SLCAE operated on multiple multi-view depth images of the 3D object. In SLCAE, FC layers were substituted by layers that were connected locally with the use of the convolution operation. Multiple AEs ere stacked where the output of the last AE was used as a final representation of the 3D object. Experiments on different datasets: SHREC'09, NTU and PSB proved the capabilities of this model. As an extension to the previous work, Leng et al. proposed a ``\textit{3D Convolutional Neural Network} (3DCNN)" to simultaneously process different 2D views of the 3D object~\cite{leng20153db}. Different views are sorted in a specific order to guarantee that all the objects' views follow the same convention while training. The proposed 3DCNN is composed of four Conv layers, three sub-sampling layers and two FC layers. The proposed network was tested for retrieval task on the same datasets used for testing~\cite{leng20153da}. However, the results showed that the later model performed better on the three datasets which implies that the previous model was able to learn more discriminative features to represent various 3D objects. 
			
			\par
\begin{figure}[h]
				\centering
				\includegraphics[width=1.0\columnwidth]{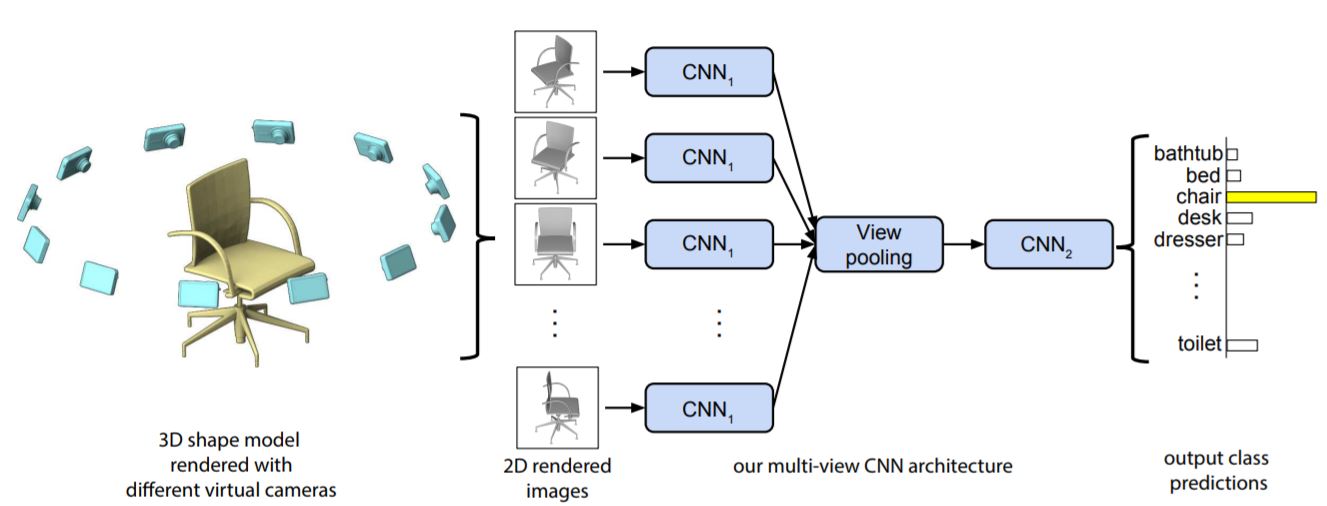}\caption{MVCNN architecture~\cite{su2015multi} applied on multi-view of 3D objects without a specific order. The figure reused from~\cite{su2015multi} with permission of authors.}
				\label{fig:MVCNN}
				\end {figure}

 A novel ``\textit{Multi-View CNN} (MVCNN)" was proposed by Su et al. in~\cite{su2015multi} for 3D object retrieval and recognition/classification tasks. In contrast to Leng's model in~\cite{leng20153db}, MVCNN processed multiple views for the 3D objects in no specific order using a view pooling layer. \ref{fig:MVCNN} shows the full architecture of the model. Two different setups to capture the 3D objects multi-views were tested. The first one rendered 12 views for the object by placing 12 equidistant virtual cameras surrounding the object while the other setup included 80 virtual views. MVCNN was pre-trained using $ImageNet1K$ dataset and fine-tuned on $ModelNet40$~\cite{wu20153d}. The proposed network has two parts, the first part is where the object's views are processed separately and the second part is where the max pooling operation is taking place across all the processed views in the view-pooling layer, resulting in a single compact representation for the whole 3D shape. In the view-pooling layer, the view with the maximal activation is the only one considered while ignoring all the other views with non-maximal activations. This means that only few views are contributing towards the final representation of the shape which causes a loss of the visual information. To overcome this problem,

			Experiments showed that the MVCNN with the max-view pooling layer outperformed ShapeNet \cite{wu20153d} on classification and retrieval tasks by a remarkable margin. In~\cite{johns2016pairwise} Johns et al. exploited multi-view data representation using CNNs by representing 3D objects under unconstrained camera trajectories with a set of 2D image pairs. The proposed method classifies each pair separately and then weight the contribution of each pair to get the final result. The VGG-M architecture was adopted in this framework consists of five Conv layers and three FC layers. The views of the 3D objects are represented as either depth images or grayscale images or both. This model outperformed MVCNN proposed by Su et al.~\cite{su2015multi} and voxel-based ShapeNet architectures~\cite{wu20153d}.

			The efficiency of multi-view DL models pushed researchers to investigate more GPU-based methods to learn multi-view 3D data features. This is what pushed Bai et al.~\cite{bai2016gift} to propose a real-time GPU-based CNN search engine for multi 2D-views of 3D objects. The proposed model called $GIFT$ utilizes two files that are inverted: the first is to accelerate process of the multi-view matching and the second one is to rank the initial results. The processed query is completed within one second. $GIFT$ was tested on a set of various datasets: \textit{ModelNet}, $PSB$, $SHREC14LSGTB$, $McGill$ and $SHREC'07$ watertight models. GIFT produced a better performance compared to the state-of-the-art methods. 
			
			The efforts to learn multi-view 3D data representations kept evolving and in~\cite{zanuttigh2017deep}, Zanuttigh and Minto proposed a multi-branch CNN for classifying 3D objects. The input to this model is rendered depth maps from different view points for the 3D object. Each CNN branch consists of five Conv layers to process one depth map producing a classification vector. The resulted classification vectors are the input to a linear classifier to identify the 3D object's category/class. The proposed model produced comparable results to the state-of-the-art. Based on the dominant sets, Want et al. in~\cite{wang2017dominant} proposed recurrent view-clustering and pooling layers . The key concept in this model is to pool similar views and recurrently cluster them to build a pooled feature vector. Then, the constructed pooled feature vectors are fed as inputs in the same layer in a recurrent training fashion in the recurrent clustering layer. Within this layer, a view similarity graph is computed whose nodes represent the feature vectors and the edges represent the similarity weights between the views. Within the constructed graph, the similarities and dissimilarities between different views are exhibited which is very effective in the 3D shape recognition task. The proposed model achieved a highly comparable results to previous methods ~\cite{wu20153d, su2015multi} as shown in Table 1 in the supplementary material. 
			Driven the advances in the multi-view DL models, Qi et al.~\cite{qi2016volumetric} provided a comparison study between multi-view DL techniques and volumetric DL techniques for the object recognition task. As part of the study, the authors proposed a $Sphere rendering$ approach for filtering multi-resolution 3D objects at multiple scales. With data augmentation, the authors managed to enhance the results of MVCNNs on $ModelNet40$. Recently, Kanezaki et at.~\cite{kanezaki2016rotationnet} achieved state-of-the-art results on both $ModelNet10$ and $ModelNet40$ in the classification problem using $RotationNet$. $RotationNet$ trains a set of multi-view images for the 3D object but doesn't require all the views at once. Instead, it allows for sequential input and updates the likelihood of the object's category accordingly.  
			
			Multi-view representation proved to perform slightly better than volumetric representation with less computational power needed. However, there are some challenges imposed with this representation. The sufficient number of views and the way they were acquired is a critical factor for representing the 3D shape. Also, the multi-view representation does not preserve the intrinsic geometric properties of the 3D shape. This is what pushed towards defining new notion of convolution operating on 3D shapes to capture their intrinsic properties.

				\subsubsection{Deep learning architectures on hybrid data representations}
				\par Some efforts towards combining various 3D data representations to exploit the advantages that each representation brings. Recently, Wang et al. proposed ~\cite{wang2015sketch}, where each 
				of the 3D objects are represented by a pair of multi-views and 2D sketches.
				The learning model is composed of Siamese CNNs where are two identical sub-convolutional networks; one for processing the multi-views input and the other one is for processing the 2D sketches. Each of these network consisted of three Conv layers where each of them
				was succeeded by a max-pooling layer and a Fully Connected Layer.
				The networks were trained separately using the 
				``\textit{Stochastic Gradient Descent} (SGD)" 
				method. The proposed model was tested on SHREC'13 dataset for 3D shape retrieval and achieved competitive results compared to previous methods. Wang et al. continued their investigations for hybrid 3D data representations and in~\cite{wang2016efficient}, the authors proposed the ``\textit{Convolutional Auto-Encoder Extreme Learning Machine} (CAE-ELM)" 3D descriptor which 
				merges
				the learning power of ConvNets, AEs and ``\textit{Extreme Learning Machine} (ELM)"~\cite{huang2006extreme}. ELM is an efficient unsupervised learning technique 
				which
				learns high-level discriminative features about the input data. ELM is faster than most of DL models~\cite{kasun2013representational} which is practical for processing large-scale 3D data. The input to the CAE-ELM architecture is two data representations: voxel data and ``\textit{Signed Distance Field data} (SDF)". Voxel data 
				explains
				the structure of the 3D object while the SDF extracts local and global features about the 3D object. CAE-ELM was tested on $ModelNet40$ and $ModelNet10$ datasets for classification tasks and achieved a superior performance compared to previous methods. CAE-ELM is considered as a hybrid approach that exploits the structure of the 3D objects in addition to 3D descriptors. 
				Ben-Shabat et al.~\cite{ben20173d} proposed
				a novel 3D point cloud representation 
				named ``\textit{3D Modified Fisher Vectors} (3DmFV)" 
				which is a 
				kind of 
				DL model that uses a hybrid data representation of the discrete structure of a grid with the continuous generalization of Fisher vectors to represent the 3D data. The hybrid input data is processed using deep ConvNet for classification and part segmentation tasks. 3DmFV has two modules: 
				the 
				initial one 
				changes the input point cloud into a 3D modified ``\textit{Fisher vector} (FV)" which can be considered as a descriptor-based representation and the second one is the DL module represented in the CNN. FV data representation 
				empowers
				the proposed framework to 
				become order, input data sample size and structure invariant.
				The network architecture is composed of an inception module~\cite{szegedy2017inception}, max-pooling layers and four FC layers on top. 
				this network has achieved better results in comparison with state-of-the-art techniques.
				
				Inspired by the performance of 3D convolution and 2D multi-view CNNs, some work examined the fusion of both representations. Hedge and Zadeh~\cite{hegde2016fusionnet}, proposed to fuse the volumetric representations (voxels) and 2D representation (2D views) for the object classification task. Authors have examined different models for processing the data for $ModelNet$ classification. The model one was a combination of both modalities (3D and 2D) while using two 3D CNNs for processing the 3D voxels and AlexNet for processing the 2D views. The other experiments were carried by processing the 3D voxels only and compare it with the performance of AlexNet on the 2D views. Experiments showed that that network that combined both modalities called $FusionNet$ performed the best. However, the multi-view network performed better than the volumetric CNNs. Although Brock et al. in ~\cite{brock2016generative} 
				has achieved a state-of-the-art results on $ModelNet$ classification,
				$FusionNet$ achieved comparable results with no need for the data augmentation or the very heavy computations needed in Brock's model. This implies that practically employing 2D and 3D representations surpass the volumetric methods with less computations.

				\subsection{Deep learning architectures on 3D non-Euclidean structured data}~\label{Deep learning architectures on 3D non-Euclidean structured data}
				The second type of 3D DL approaches is the non-Euclidean approaches that try to extend the DL concept to geometric data. However, the nature of the data is imposing challenges on how to perform the main DL operations such as convolution. A number of architectures that tried to extend DL to the 3D geometric domain were proposed. Some of these architectures addresses 
				``3D point clouds", in order to learn the geometry of a 3D shape
				and use it for modeling tasks. Results encouraged researchers to leverage the surface information provided in 3D meshes where the connectivity between vertices can be exploited to define local pseudo-coordinates to perform a convolution-like operation on 3D meshes. At the same time, some efforts were directed towards investigating graph-based 3D data attempting to leverage the spectral properties of graphs to define intrinsic descriptors to be used in the DL framework. In this section, we will go over the latest innovations in applying DL on non-Euclidean 3D data.
				\subsubsection{Point clouds} \label{Point clouds}
				\par Point clouds provide an expressive, homogeneous and compact representation of the 3D surface geometry without the combinatorial irregularities and complexities of meshes. That is why point clouds are easy to learn from. However, processing point clouds is tricky due to their dual nature. Point clouds can be seen as a  Euclidean-structured data locally when considering a point with respect to its neighborhood (a subset of points) such that the interaction among the points forms a Euclidean space with a distance metric which is invariant to transformations 
				like
				translation, rotation. However, considering the global structure of the point cloud, it is an unordered set of points with no specific order which imposes the irregular non-Euclidean nature on the global structure of the data.    
				\par Some recent works have considered point clouds as a collection of sets with different sizes. Vinyals et al. in~\cite{vinyals2015order} use a read-process-write network for processing point sets to show the network ability to learn how to sort numbers. This was a direct application of DL on an unordered set for the Natural Language Processing (NLP) application. Inspired by this work, Ravanbakhsh et al.~\cite{ravanbakhsh2016deep}, proposed what they called the permutation equivariant layer within a supervised and semi-supervised settings. This layer is obtained by parameter-sharing to learn the permutation invariance as well as rigid transformations across the data. This network
				performed 3D classification and MNIST digit addition.
				Although this network is relatively simple, 
				it didn't perform well on the 3D classification task on \textit{ModelNet} dataset. A deeper version of this model was extended in~\cite{zaheer2017deep} where the proposed $DeepSet$ framework 
				produced better results than the state-of-the-art methods in 3D classification job on $ModelNet$ dataset.
				This is due to the permutation invariance property that the permutation equivarant layer is bringing to the previous models.~\cite{qi2016pointnet} also used a similar layer with a major difference as the permutation equivariant layer is max-normalized.
				
				\par $PointNet$~\cite{qi2017pointnet} is the pioneer in making a direct use of the point cloud as an input where each of its points is represented using the $(x,y,z)$ coordinates. As a pre-processing step, 
				feature transformation and inputs are feeded into the $PointNet$ architecture.
				$PointNet$ is composed of three main modules: a ``Spatial Transformer Network (STN)" module, an RNN module and a simple symmetric function that aggregates all the information from each point in the point cloud. The STN canonicalizes the data before feeding them to the RNN, i.e., process all the data into one canonical form, and learns the key points of the point cloud which approximately corresponds to the skeleton of the 3D object. Then comes the RNN module which learns the point cloud 
				like a sequential signal of points and while training this model with some randomly permuted sequence, this RNN becomes invariant to the sequence of the input order of the point cloud's point.
				Lastly, the network aggregates all the resulted point features using the max-pooling operation which is also permutation invariant. $PointNet$ proved that it is robust against partial data and input perturbation. It was tested on classification and segmentation tasks where it proved to produce results comparable to the state-of-the-art as shown the supplementary material, Table 1.
				
				\par Despite the competitive results achieved by $PointNet$, it 
				is not able to take full advantage of the
				point's local structure to capture the detailed fine-grained patterns because of aggregating all the point features together. To address this point, \textit{PointNet++}~\cite{qi2017pointnet++} builds on $PointNet$ by recursively applying it
				to a nested partition of the input point sets.
				Despite capturing more features,  the resulted architecture is complicated which increases the size of the higher features and the computational time.
				
				\par Instead of operating directly on the point clouds structure, Kd-Networks by Klokov et al.~\cite{klokov2017escape} proposes to impose a kd-tree structure of the input point cloud to be used for learning the shared weights across the points of the tree. Kd-tree is a feed-forward network that has learnable parameters associated with the weights of the nodes in the tree. This model was tested for shape classification, shape retrieval and shape part segmentation producing competitive results. Following the same concept of not working directly on the point cloud structure, Roveri et al.~\cite{roveri2018network} proposed to extract a set of 2D depth maps from different views of the point cloud and process them using Residual Nets ($ResNet50$)~\cite{He2015}. The proposed framework contains 3 modules. The initial module is responsible of learning $k$ directional views of the input point cloud to generate the depth maps accordingly in the second module. The third and last module is processing the generated $k$ depth maps for object classification. The innovation of this framework is mainly focus on automatically transforming un-ordered point clouds to informative 2D depth maps without the need to adapt the network module to account for permutation invariance and different transformations of the input data. 
				
				In recent past, few articles have reported their work on unsupervised learning over the point clouds.
				In $FoldingNet$~\cite{yang2018foldingnet}, Yang et al. proposed to use AE for modelling different 3D objects represented as point clouds by a novel folding-based decoder that deforms a 2D canonical grid into the underlying surface of the 3D point cloud. $FoldingNet$ is able to learn how to generate cuts on the 2D grid to create 3D surfaces and generalize to some intra-class variations of the same 3D object class. An SVM was used on top of this $FoldingNet$ to be used for 3D classification where it proved to perform well with the learned discriminative representation for different 3D objects. \textit{FoldingNet} achieved high classification accuracy on $ModelNet40$. Another unsupervised model was proposed by Li et al. called \textit{SO-Net}~\cite{li2018so}. \textit{SO-Net} is a permutation invariant network that can tolerate unordered point clouds inputs. \textit{SO-Net} builds that spatial distribution of the points in the point cloud using ``\textit{Self-Organizing Maps} (SOMs)". Then, a hierarchical feature extraction on the points of the point cloud and the SOM nodes is employed 
				which results in a singular feature vector which represents the entire point cloud
				Local feature aggregation happens according to an adjustable receptive field where the overlap is controlled to get more effective features. \textit{SO-Net} was tested on classification and segmentation tasks producing promising results highly comparable with the state-of-the-art techniques as shown in Table 1 of the supplementary material.

				As noticed in all the previously proposed methods, the main problem in processing point clouds is the un-ordered structure of this representation where researchers are trying to make the learning process invariant to the order of the point cloud. Most of these methods resorted to clustering techniques to opt for similar points and process them together.

				\subsubsection {Graphs and meshes} \label{DL_graphsAndMeshes}
				
				The ideal representation for graphs and meshes is the one that can capture all the intrinsic structure of the object and also can be learned with the gradient descent methods. This is due to their stability and frequent usage in the CNNs. However, learning such irregular representations is a challenging task due to the structural properties of these representations. Motivated by the success of CNNs in a broad range of computer vision tasks, recent research efforts were directed towards generalizing CNNs to such irregular structures. Analyzing the properties of such data shows that meshes can be converted to graphs as discussed in Section~\ref{dataRep_3D meshes and graphs.}. Hence, the models proposed for graphs can be employed on mesh-structured data but not vice versa. Most of the existing work addresses the graph-structured data explicitly with some few works were tailored towards mesh representations. We herein overview recent works on each representation providing a broad classification for the existing methods based on the used approach. 
				
				{\bf Graphs:} Studying the structural properties of both graphs and meshes suggests that the proposed learning methods for graphs are also applicable on meshes. Existing methods for Graph Convolutional Neural Networks (GCNN) can be broadly categorized into two main directions: spectral filtering methods and spatial filtering methods. Here we discuss the underlying concept behind each method and overview the work done in each direction. The distinction between both directions is in how the filtering is employed and how the locally processed information is combined. 
				
				\begin{itemize}                          
					
					\item{\it Spectral filtering methods.} The notion of spectral convolution on graph-structured data was introduced by Bruna et al. in~\cite{bruna2013spectral} where the authors proposed Spectral CNN (SCNN) operating on graphs. The foundation of the spectral filtering methods is to use the spectral eigen-decomposition of the graph Laplacian to define a convolution-like operator. This redefines the convolution operation in the spectral domain where the main two core stones are analogous: the patches of the signal in the Euclidean domain correspond to the functions defined on the graph nodes e.g. features, mapped to the spectral domain by projecting on the eigenvectors of the graph Laplacian. The filtering operation itself happens in the Euclidean domain and corresponds to scaling the signals in the eigenbasis. This definition implies that convolution is a linear operator that commutes with the Laplacian operator~\cite{bronstein2017geometric}. Despite the innovation aspect of Bruna's model, it has serious limitations due to being basis dependent and computationally expensive. Being basis-dependent means that if the spectral filter's coefficients were learned with respect to a specific basis, applying the learned coefficients on another domain with another basis will produce very different results as illustrated in~\cite{bronstein2017geometric}. The other limitation of being computationally costly arises from the fact that spectral filtering is a non-local operation that involves data across the whole graph besides that the graph Laplacian is expensive to compute. This constitutes a computational burden towards generalizing to other bases and processing large-scale graphs.  
					
					The work in~\cite{kovnatsky2013coupled} addressed the basis dependency problem by constructing a compatible orthogonal basis across various domains through a joint diagonalization. However, this required a prior knowledge about the correspondence between the domains. 
					For some applications like social networks,
					this is a valid assumption because the correspondence can be easily computed between two time instances in which new edges and vertices have been added. However, applying this on meshes is rather unreasonable because finding correspondence between two meshes is challenging task on its own. Therefore, assuming the knowledge of correspondence between domains in such case is unrealistic~\cite{bronstein2017geometric}. Since the non-local nature of the spectral filtering and the need to involve all the graph data in the processing, recent works proposed the idea of approximation to produce local spectral filters~\cite{defferrard2016convolutional}~\cite{kipf2016semi}. These methods propose to represent the filters via a polynomial expansion instead of directly operating on the spectral domain. Defferrard et al. in ~\cite{defferrard2016convolutional} performed local spectral filtering on graphs by using Chebyshev polynomials 
					in order to approximate the spectral graph filters.
					The features yielding from the convolution operation are then coarsened using the graph pooling operation. Kipf and Welling~\cite{kipf2016semi} simplified the polynomial approximation proposed in~\cite{defferrard2016convolutional} and used a first-order linear approximation of the graph spectral filters to produce local spectral filters which are then employed in a two-layer GCNN. Each of these two layers uses the local spectral filters and aggregates the information from the immediate neighbourhood of the vertices. Note that the filters proposed in~\cite{defferrard2016convolutional} and ~\cite{kipf2016semi} are employed on r- or 1-hop neighbourhood of the graph returns these constructions into the spatial domain.
					
					Driven by the success of the local spectral filtering models, Wang et al.~\cite{wang2018local} proposed to 
					take the advantages of the power of spectral GCNNs in the 
					pointNet++ framework~\cite{qi2017pointnet++} to process unordered point clouds. This model fuses the innovation of the pointNet++ framework with local spectral filtering while addressing two shortcomings of these models independently. Therefore, instead of processing each point independently in the point clouds as proposed in pointNet++, this model uses spectral filtering as a learning technique to 
					change the structural information of each points' neighborhood. Moreover, rather than 
					using the greedy winner-takes all method in the graph max pooling operation, this method adopts a recursive pooling and clustering strategy. Unlike the previous spectral filtering methods, this method 
					does not require any pre-computation and it is trainable by an end-to-end manner which 
					allows building the graph dynamically and computing the graph Laplacian and the pooling hierarchy on the fly unlike~\cite{bruna2013spectral,defferrard2016convolutional,kipf2016semi}. 
					This method have been able to achieve better recognition results than the existing state-of-the-art techniques on diverse datasets as shown in Table 1 of the supplementary material. 
					\item{\it Spatial filtering methods.} The concept of graph spatial filtering started in~\cite{scarselli2009graph} when GNNs were first proposed as an attempt to generalize DL models to graphs. GNNs are simple constructions that try to generalize the notion of spatial filtering on graphs via the weights of the graph. GNNs are composed of multiple layers where each layer is a linear combination of graph high-pass and low-pass operators. This formulation suggests that learning the graph features is dependent on each vertex's neighborhood. Similar to Euclidean CNNs, a non-linear function is applied to all the nodes of the graph where the choice of this function varies depending on the task. Varying the nature of the vertex non-linear function lead to rich architectures~\cite{li2015gated,sukhbaatar2016learning,duvenaud2015convolutional,chang2016compositional,battaglia2016interaction}. Also, analogous to CNNs, pooling operation can be employed on graph-structured data by graph coarsening. Graph pooling layers can be performed by interleaving the graph learning layers. In comparison with the spectral graph filtering, spatial filtering methods have two key points which distinguish them from spectral methods. Spatial methods aggregate the feature vectors from the neighborhood nodes directly based on the graph topology considering the spatial structure of the input graph. The aggregated features are then summarized via an additional operation. 
					The GNN framework presented in~\cite{scarselli2009graph,gori2005new}, proposed to embed each vertex in the graph into a Euclidean space with an RNN. Instead of using the recursive connections in the RNN, the authors used a simple diffusion function for their transition function, propagating the node representation repeatedly until it is stable and fixed. The resulting node representations are considered as the features for classification and regression problems. The repeated propagation for node features in this framework constitutes a computational burden which is alleviated in the work proposed by Li et al.~\cite{li2015gated}. 
					They have 
					proposed a variant of the previous model which uses the gated recurrent units to perform the state updates to learn the optimal graph representation. Bruna et al. in~\cite{bruna2013spectral} imposed the spatial local receptive field on GNN to produce their local spatial formulation of GNN. The main idea behind the local receptive field is to 
					decrease
					the number of the learned parameters by grouping similar features based on a similarity measure~\cite{coates2011selecting,gregor2010emergence}. In~\cite{bruna2013spectral}, the authors used this concept to compute a multi-scale clustering of the graph to be fed to the pooling layer afterwards. This model imposes locality on the processed features and reduces the amount of processed parameters. However, it doesn't perform any weight sharing similar to 2D CNNs. Niepert et al.in~\cite{niepert2016learning} performs spatial graph convolution in a simple way by converting the graph locally into sequences and feeding these sequences into a 1D CNN. This method is simple but requires an explicit definition for the nodes orders of the graphs in a pre-processing step. 
					In~\cite{venkatakrishnan2018graph2seq}, the authors provided a detailed study proving that spectral methods and spatial methods are mathematically equivalent in terms of their representation capabilities. The difference resides in how the convolution and the aggregation of the learned features are performed. Depending on the task, the architecture of the GCNN (spectral or spatial) is formed where the convolution layers may be interleaved with coarsening and pooling layers to summarize the output of the convolution filters for a compact representation of the graph. This is crucial in classification applications where the output is only one class inferred from the learned features~\cite{bruna2013spectral}. Some other applications require a decision per node such as community detection. A common practice in such cases is to have multiple convolution layers that compute the graph representations at the node level ~\cite{khalil2017learning,nowak2017note,bruna2017community}. All these GCNNs are end-to-end differentiable methods that can be trained in supervised, semi-supervised or reinforcement learning techniques.
					
				\end{itemize}                          
				{\bf Meshes:}
				On the Euclidean domain, the convolution operation is performed by passing a template at each point on the spatial domain and recording the correlation between the templates using the function that is defined at this point. 
				This is feasible due to the shift-invariance property on the Euclidean domain. Howeverm, unfortunately, this is not directly applicable on meshes 
				because there is a
				lack of the shift-invariance property. This is what pushed towards defining local patches that represent the 3D surface in a way that allows performing convolution. However, due to the lack of global parametrization on non-Euclidean data, these patches are defined in a local system of coordinates locally meaning that these patches are also position-dependent. Recently, various non-Euclidean CNNs frameworks were proposed. The main schema used by these frameworks is very similar except for how the patches are defined mostly. The local patches are defined either by handcrafting them or depending on the connectivity of the vertices while using the features of the 1-hop neighborhood as the patch directly~\cite{fey2017splinecnn}. The convolution employed in such frameworks is very similar to the classical 2D convolution where it is basically an element-wise multiplication between the convolution filter and the patch and summing up the results. This is because the patches extracted by such frameworks boils down the representation into 2D where the classical convolution can be employed. 
				
				Geodesic CNN~\cite{masci2015geodesic} was introduced as a generalization of classical CNNs to triangular meshes. The main idea of this approach is to construct local patches in local polar coordinates. The values of the functions around each vertex in the mesh are mapped into local polar coordinates using the patch operator. This defines the patches where the geodesic convolution is employed. Geodesic convolution follows the idea of multiplication by a template. However, the convolution filters in this framework are subjected to some arbitrary rotations due to the angular coordinate ambiguity~\cite{masci2015geodesic}. This method opened the door for new innovations on extending CNN paradigm to triangular meshes. However, this framework suffers from multiple drawbacks. First, it can only be applied on triangular meshes where it is sensitive to the triangulation of the mesh and it might fail if the mesh is extremely irregular. Second, the radius of the constructed geodesic patch has to be small with respect to the injectivity radius of the actual shape to guarantee that the resulted patch is topologically a disk. Third, the rotations employed on the convolution filters make the framework computationally expensive which limits the usage of such a framework. Anisotropic CNN (ACNN)~\cite{boscaini2016learning} was proposed to overcome some of the limitations in the geodesic CNN. Compared to the geodesic CNN, the ACNN framework is not limited to triangular meshes and can be also applied to graphs. Also, the construction of the local patches is simpler and is independent on the injectivity radius of the mesh. ACNN uses the concept of spectral filtering where the spatial information is also incorporated by a weighting function to extract a local function defined on the meshes. The learnt spectral filters are applied to the eigenvalues of the anisotropic Laplacian Beltrami Operator (LBO) and the anisotropic heat kernels act as a spatial weighting functions for the convolution filters. This method has shown a very
				good performance for local correspondence tasks. Rather than using a fixed kernel construction as in the previous models, Monti et al.~\cite{monti2017geometric} proposed $MoNet$ as a general construction of patches. The authors proposed to define a local system of coordinates of pseudo-coordinates around each vertex with weight functions. On these coordinates, a set of parametric kernels are applied on these pseudo-coordinates to define the weighting functions at each vertex. That is why the previous methods~\cite{masci2015geodesic,boscaini2016learning} can be considered as specific instances of \textit{MoNet}. 
				Some recent work has been proposed to eliminate the need to explicitly define the local patches on the graphs or meshes such as SplineCNN~\cite{fey2017splinecnn}. SplineCNN is a convolutional framework that can be employed on directed graphs of any dimensionality. Hence, it can also be applied on meshes. Instead of defining the local patches by a charting-based method like the previous methods, SplineCNN uses the 1-hop neighborhood ring features of the graph as the patch where the convolutional filter can operate. The convolutional filter itself is a spatial continuous filter based on B-Spline basis functions that have local support. This framework produces state-of-the-art results on the correspondence task while being computationally very efficient. This is due to the local support of the B-Spline basis which makes the computational time independent of the kernel size.

				\section{Analysis and discussions} \label{analysis}
				DL paradigms are successfully architect and deployed to various 3D data
				representations as discussed in the previous sections. Several approaches have been proposed. We herein discuss the main 3D datasets and their exploitation in various 3D computer vision tasks. Also, we present DL advances in three main tasks; 3D recognition/classification, retrieval and correspondence. 
				
				
				\subsection{3D Datasets} \label{3D datasets}                                    
				We overview below the most recent 3D datasets. There are two main categories of data used by the research community: real-world datasets and synthetic data rendered from CAD models. It is preferable to use the real-world data; however, real data is expensive to collect and usually suffers from noise and occlusion problems. In contrast, synthetic data can produce a huge amount of clean data with limited modelling problems. While this can be seen advantageous, it is quite limiting to the generalization ability of the learned model to real-world test data. It is also important to note that most 3D datasets are smaller than large 2D datasets such as, $ImageNet$~\cite{imagenet_cvpr09}. However, there are some recent exceptions as described below.
				
				\textit{ModelNet}~\cite{wu20153d} is the most commonly used dataset for 3D object recognition and classification. It contains roughly 130k annotated CAD models on 662 distinct categories. This dataset was collected using online search engines by querying for each of the categories. Then, the data were manually annotated. ModelNet provides the 3D geometry of the shape without any information about the texture. ModelNet has two subsets: \textit{ModelNet10} and \textit{ModelNet40}. These subsets are used in most of the recently published work as shown in Table 1 and Table 2 in the supplementary material. In Section~\ref{3D recognition}, we provide an extensive analysis of the methods employed on the ModelNet dataset for recognition and retrieval tasks highlighting the evolution of DL methods for learning such data.

				\textit{SUNCG}~\cite{song2017semantic} contains about 400K of full room models. This dataset is synthetic, however, each of these models was validated to be realistic and it was processed to be annotated with labelled object models. This dataset is important for learning the ``scene-object" relationship and to fine-tune real-world data for scene understanding tasks. \textit{SceneNet}~\cite{handa2016scenenet} is also an RGB-D dataset that uses synthetic indoor rooms. This dataset contains about 5M scenes that are randomly sampled from a distribution to reflect the real world. However, not all the generated scenes are realistic and in practice, some of them are highly unrealistic. Still, this dataset can be used for fine-tuning and pre-training. In contrast, \textit{ScanNet}~\cite{dai2017scannet} is a very rich dataset for real-world scenes. 
				It is an annotated dataset which is labelled with some semantic segmentation, camera orientation and the 3D information that is gathered from 3D video sequences of real indoor scenes. 
				It includes 2.5M views, which allows for training directly without pre-training on other datasets, as it is the case with different datasets. 
				
				In addition, datasets for 3D meshes are available for the 3D computer vision community. Most of the 3D meshes datasets are for 3D objects, body models or face data. The \textit{TOSCA}~\cite{bronstein2008numerical} dataset provides high-resolution 3D synthetic meshes for non-rigid shapes. It contains a total of 80 objects in various poses. Objects within the same category have the same number of vertices and the same triangulation connectivity. \textit{TOSCA}~\cite{bronstein2008numerical} provides artistic deformations on the meshes to simulate the real-world deformations of real scans. \textit{SHREC}~\cite{bronstein2010shrec} adds a variety of artificial noise and artistic deformations on \textit{TOSCA} scans. However, the artificial noise and deformations are not realistic and can’t generalize to new unseen real-world data which is a requirement for practical solutions. \textit{FAUST}~\cite{Bogo:CVPR:2014} dataset, however, provides 300 real-scans of 10 people in various poses. The \textit{3DBodyTex} dataset~\cite{saint20183dbodytex} is recently proposed with 200 real 3D body scans with high-resolution texture. \textit{3DBodyTex} is a registered dataset with the landmarks available for 3D human body models.
				
				The series of the \textit{BU datasets} is very popular for 3D faces under various expressions. \textit{BU-3DFE}~\cite{yin20063d} is a static dataset which has 100 subjects (56 female and 44 male) of different ages and races. Each subject has in addition to the neutral face, six expressions as (happiness, sadness, anger, disgust, fear and surprise) of different intensities. There is 25 meshes for each subject in total, resulting in 2500 3D facial expressions dataset. Another very popular dataset is \textit{BU-4DFE}~\cite{yin2008high}, which is a dynamic facial expression dataset that has 101 subjects in total (58 female and 43 male). Similar to the \textit{BU-3DFE}, each subject has six expressions. Other 3D faces datasets were available as well like the \textit{BP4D-Spontanous}~\cite{zhang2014bp4d} and \textit{BP4D+}~\cite{zhang2016multimodal}.

				\subsection{3D Computer vision tasks}    \label{3D Computer Vision tasks}        
				After the huge success of DL approaches in various computer vision tasks in the 2D domain, 
				DL methods have gained more popularity as they are producing some remarkable performances on different tasks.
				Here, we overview 3D DL advances on the 3D object recognition/classification, retrieval and correspondence tasks.

				\subsubsection{3D recognition/classification}    \label{3D recognition}

				The tasks of 3D recognition/classification is fundamental in computer vision. Given a 3D shape, the goal is to identify the class to which this shape belongs (3D classification) or given a 3D scene, recognize different 3D shapes in the scene along with their positions (3D recognition and localization).
				There is an active research to exploit Deep Neural Networks (DNNs) for 3D object recognition/classification. Existing approaches can be classified according to their input to learn the task as shown in Table 1 and Table 2 in the supplementary material. Multi-view approaches perform the task of classification using the learned features after applying 2D CNNs on each view. Generally, the multi-view methods outperform the other methods; however, there are still some unsolved drawbacks. For example, the recent Multi-View CNN (MVCNN)~\cite{su2015multi} applies a max-pooling operation on the features of each view to produce global features that represent the 3D object. The max-pooling ignores the non-maximal activation and only keeps the maximal ones from a specific view which results into losing some visual cues~\cite{su2015multi}. Yu et al.~\cite{yu2018multi} tried to incorporate the other views by a sum-pool operation. However, it performed worse than the max-pooling. 
				
				Considering the recognition and classification tasks,~\cite{wang2017dominant} improved the discrimination between objects by applying a recurrent clustering and pooling strategy that increases the likelihood of variations in the aggregated activation output of the multi-views. Their objective was to capture the subtle changes in the activation function space. Similarly, $GIFT$~\cite{bai2016gift} extracts the features from each view but does not apply any pooling. It matches views to find the similarity between two 3D objects. It counts the best matched views only, however the greedy selection of the best matched view may discard useful information.  Another framework called \textit{Group View CNN} (GVCNN)~\cite{Feng2018multi} contains a hierarchical architecture of content descriptions from the view level, group of views level and the shape level. The framework defines the groups based on their discrimination power and the weights are adapted accordingly. Another very recent work~\cite{yu2018multi} proposes \textit{a Multi-view Harmonized Bilinear Network} (MHBN) to improve the similarity level between two 3D objects by utilizing patch features rather than view features.
				
				Volumetric approaches classify 3D objects by using directly the shape voxels. They mitigate the challenge of having orderless points by voxelizing the input point cloud like \textit{3D ShapeNets}~\cite{Wu2015volumetric}, volumetric CNNs~\cite{qi2016volumetric}, $OctNet$~\cite{Riegler2017volumetric} and VRN Ensemble~\cite{brock2016generative}. Although VRN Ensemble outperforms the multi-view methods, this performance is thanks to the model ensemble and their advanced base model as it ensembles five ResNet models and one Inception model while most of the existing multi-view methods rely on a single VGG-M model. Generally, these methods are not as accurate as the multi-view methods. Due to data sparsity and heavy 3D convolution computations, they are heavily constrained by their resolution.
				
				$Pointset$ (Non-Euclidean) approaches classify directly the unordered point sets in order to address the sparsity problem found in volumetric methods as proposed in $PointNet$~\cite{qi2017pointnet}. For each point, $PointNet$ learns a spatial encoding and aggregates all the features to a global representation. \textit{PointNet++}~\cite{qi2017pointnet++} improves $PointNet$ by utilizing local structures formed by the metric space. The points are partitioned into local regions that overlap by the distance metric of their space. The features are then extracted in a hierarchical fine to coarse process. At the same time of $PointNet$, \textit{Kd-networks}~\cite{klokov2017escape} were proposed, it recognizes 3D models by performing multiplicative transformations and sharing their parameters given the point clouds subdivisions imposed by \textit{kd-trees}.
				
				Given any of the existing input modalities, a set of descriptors can be directly learned. The 3D representations can be also re-defined as a set of 2D geometric projections then a set of descriptors will be extracted. There is a direction to make the best out of all worlds by combining different input modalities for efficient feature extraction. A good example is the work of~\cite{bu2017hybrid} where their scheme consists of a view-based feature learning, geometry-based feature learning from volumetric representations and a modality feature fusion in which the aforementioned learnt features were associated through a Restricted Boltzman Machine (RBM).
				
\begin{figure*}[h!]
					\centering	\includegraphics[width=1.0\columnwidth]{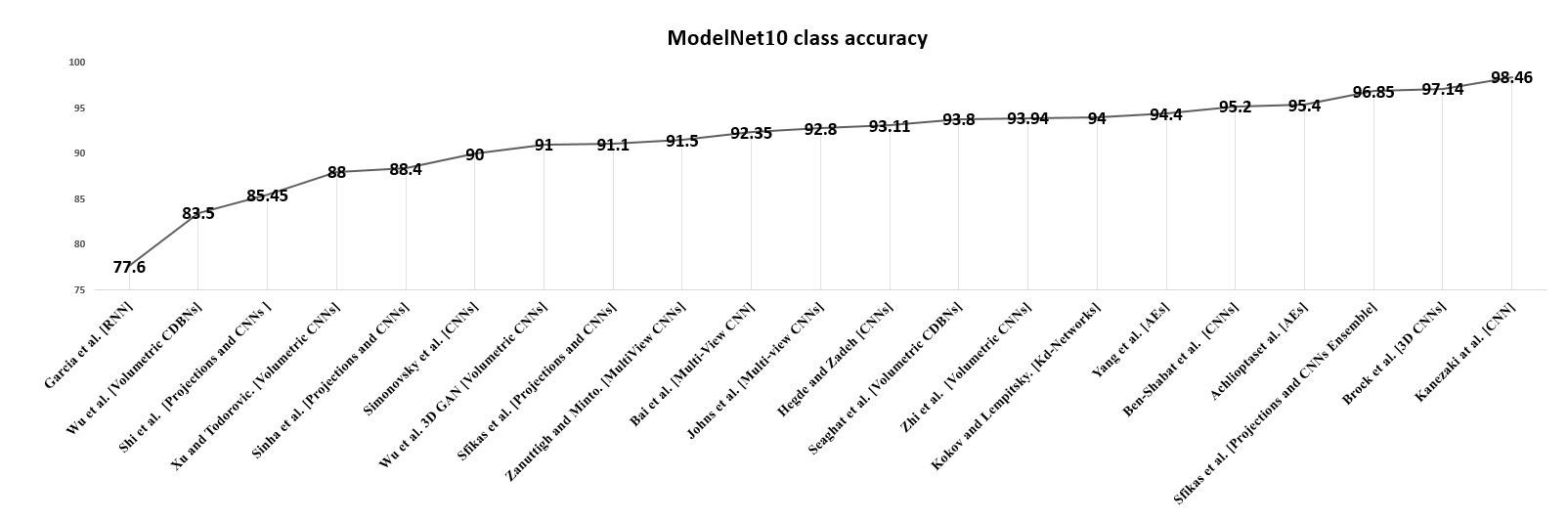}\caption{Class accuracy improvements on ModelNet10 dataset for classification/recognition tasks.}
					\label{fig:ModelNet10}
					\end {figure*}

\begin{figure*}[!t]
						\centering
						\includegraphics[ width=1.0\columnwidth]{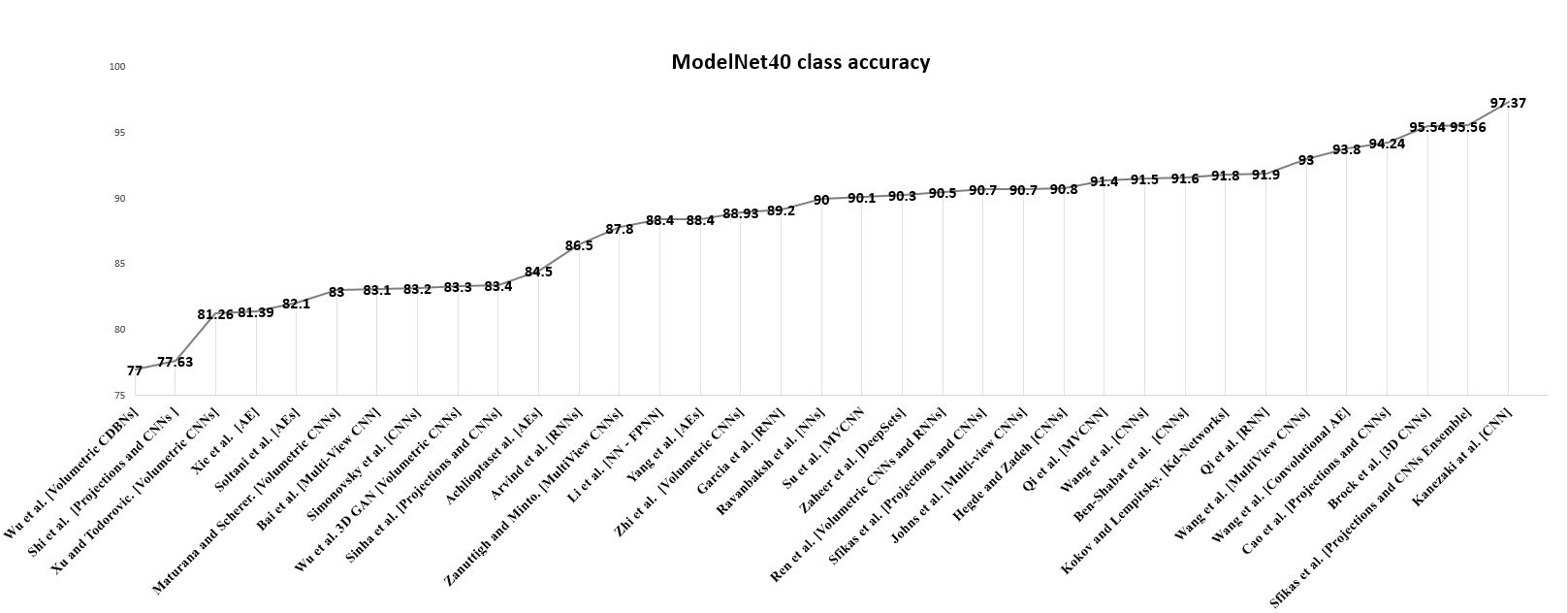}\caption{Class accuracy improvements on ModelNet40 dataset for classification/recognition tasks.}
						\label{fig:ModelNet40}
						\end {figure*}

				A comprehensive comparison between the recent state-of-the-art-methods is given in Table 1 in the supplementary material and compared based on experiments on $ModelNet10$ and $ModelNet40$ datasets in Fig.~\ref{fig:ModelNet10} and Fig.~\ref{fig:ModelNet40} respectively. Results shown in both figures highlight the power of the multi-view techniques achieving the state-of-the-art on the classification task on both datasets, $ModelNet10$ (98.46\%) and $ModelNet40$ (97.37\%). Also, it shows the competition between such methods and volumetric methods such as Brock et al.~\cite{brock2016generative} achieving very competitive results to the multi-view ($ModelNet10$ (97.14\%) and $ModelNet40$ (95.54\%)), but with a more complex architecture and a serious need for data augmentation due to the complexity of the model.

						\subsubsection{3D retrieval}    \label{3D retrieval}
						
						3D object retrieval is another fundamental task in shape analysis. 
						The target here is to find the most similar 3D object from a database to match with the tested one.
						However, in the literature, most DNNs focus on leveraging the discrimination strength of these networks for the classification and recognition tasks. There are less DNNs specifically designed for 3D shape retrieval, see Table 2 in the supplementary material. Conceptually, the input processing is similar to the classification and recognition tasks. Generally, 3D object retrieval approaches are similar to other image or object retrieval methods where several loss functions are trained to learn an embedding space to get elements closer to each other. A very recent work~\cite{He2018multi} train the center loss~\cite{wen2016centerloss} and triplet loss~\cite{Schroff2015tripletloss} specifically for the distance measure which superpasses the state-of-the-art on $ModelNet40$ and $ShapeNet$. Multi-view methods usually outperform other methods in terms of retrieval accuracy. A comprehensive list of the recent 3D object retrieval state-of-the-art methods is given in Table 2 in the supplementary material.


						
						\subsubsection{3D Correspondence} \label{coress_sec}
						The goal of 3D correspondence is to predict the mapping between a set of vertices of a test mesh and a reference or template mesh. There are two types of correspondences; sparse and dense correspondence. Sparse correspondence means that only a subset of the vertices of the test mesh are mapped to the reference mesh. However, in dense correspondence, all the vertices of the test mesh are mapped to the reference mesh.

						Several works~\cite{fey2017splinecnn,verma2018feastnet} perform 3D shape correspondence on the registered meshes of 
						\textit{FAUST} dataset~\cite{Bogo:CVPR:2014}. Only registered meshes of the dataset are considered because they are in dense point-to-point correspondence, providing the ground truth. There are in total 100 registered meshes of 10 people in 10 different poses each.
						Since the meshes are registrations of a template mesh, the number of vertices is fixed and the connectivity pattern is identical.
						In previous works, the task is cast as a classification problem. The model has to map $N$ input vertices to the corresponding vertices on the template mesh. This is achieved with a one-hot encoding of the target vertex for each input vertex.
						
						In the reported experiments of~\cite{fey2017splinecnn}, \textit{FAUST} dataset
						is divided into 80 meshes for training and 20 for testing. 
						was divided into $80$ and $20$ meshes for training and testing respectively.
						The results show very high accuracy in the correspondence prediction. Those results are very accurate, but seem to mostly reflect the simplicity of the experiment. In fact, the meshes of the dataset all have the same topology, i\@.e\@., the vertices are in the same order and the connectivity pattern is the same. Moreover, because of the registration, the vertices are already in dense one-to-one correspondence.
						There is thus no ambiguity in the possible assignments with neighboring vertices. This makes the correspondence task simple because the mapping from input vertices to ouput vertices is a trivial copy rather than learning the actual topology of the 3D mesh. 
						
						For assessing the performance of a sample state-of-the-art technique for the correspondnce task, we experimented on \textit{SplineCNN}~\cite{fey2017splinecnn}. We test the pre-trained model of \textit{SplineCNN} on different test data to validate the performance under different conditions. We have three different experiments: 1) Data with the same topology varying in the shape, pose and geometry generated from the \textit{Skinned Multi-Person Linear Model} (SMPL) model~\cite{SMPL:2015}. 2) Test data from \textit{FAUST} dataset with synthetic noise of different intensities. 
						3) Various unclothed full human body scans from the 3DBodyTex dataset~\cite{saint20183dbodytex} and some additional clothed scans acquired in a similar setup.
						
						\begin{figure*}[b!]
							\centering
							\includegraphics[width=1.0\columnwidth]{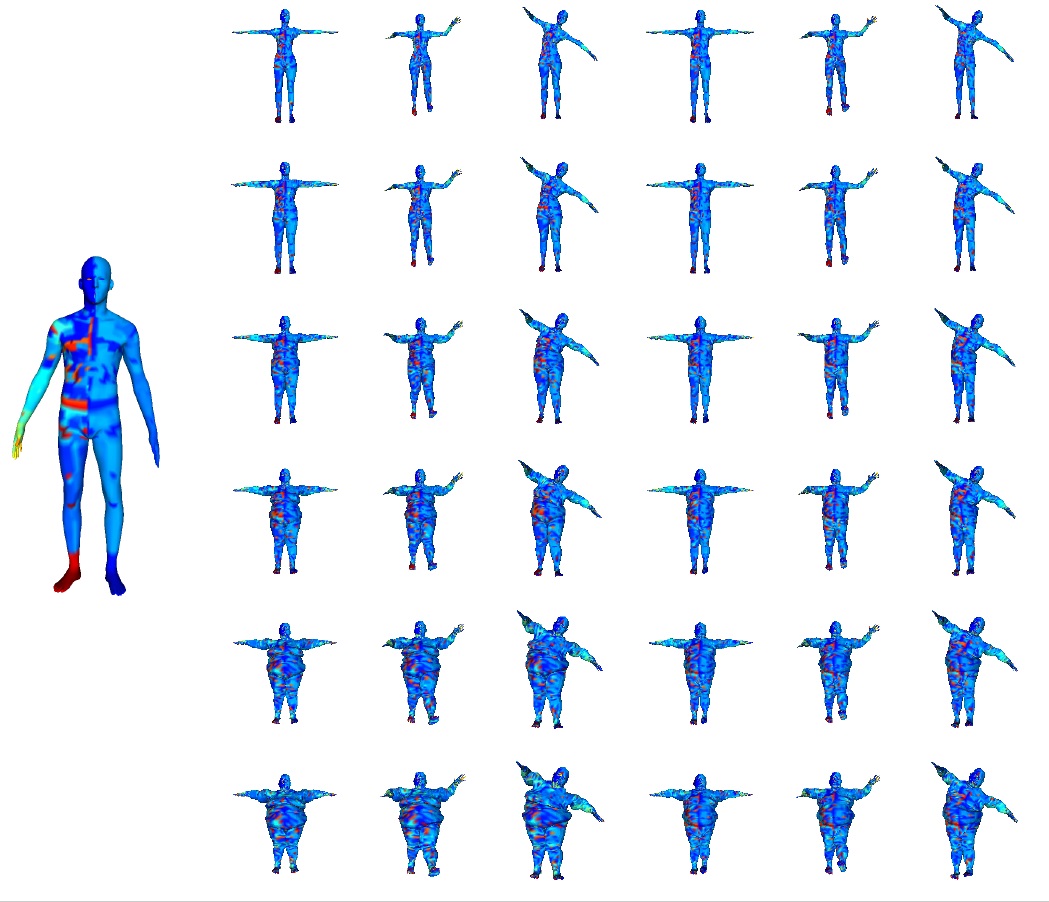}\caption{Correspondence results of testing the pre-trained model of \textit{SplineCNN}~\cite{fey2017splinecnn} on meshes generated from SMPL model~\cite{SMPL:2015}}
							\label{fig:SMPL}	
							\end {figure*}
							
							In the fist testcase, we generate different 3D meshes varying in both the shape and pose considering both genders, male and female, using the SMPL model~\cite{SMPL:2015}. The generated meshes have the same topology as the \textit{FAUST} dataset, i\@.e\@., the same number of verticies and connectivity. The generated shapes vary in size from thin to fat which in turn varies the geometry of the generated subjects. 
							We have used a pre-trained model such as \textit{SplineCNN} to test he generated data for the correspondence task. 
							Each vertex in the test subject is assigned to a specific vertex in the reference mesh and then each vertex is assigned a specific color to visualize the results.

							In Fig.~\ref{fig:SMPL}, the reference mesh is depicted on the left and all the correspondence results on the generated test data are on the right. As shown in Fig.~\ref{fig:SMPL}, the model is not able to generalize to new unseen but very similar data and does not report as good results as those reported in~\cite{fey2017splinecnn}. This performance is noticeable even on simple poses like the $T$ pose, shown in the first and fourth columns in Fig.~\ref{fig:SMPL}, which is actually included in the \textit{FAUST} dataset. Also, the results show that varying the size of the shape has the largest effect on the results more than varying the pose, as there are more wrongly colored faces in comparison with the reference mesh. However, changing the pose or the shape (male or female) does not affect the performance much.

							\begin{figure*}[b!]
								\centering	\includegraphics[width=1.0\columnwidth]{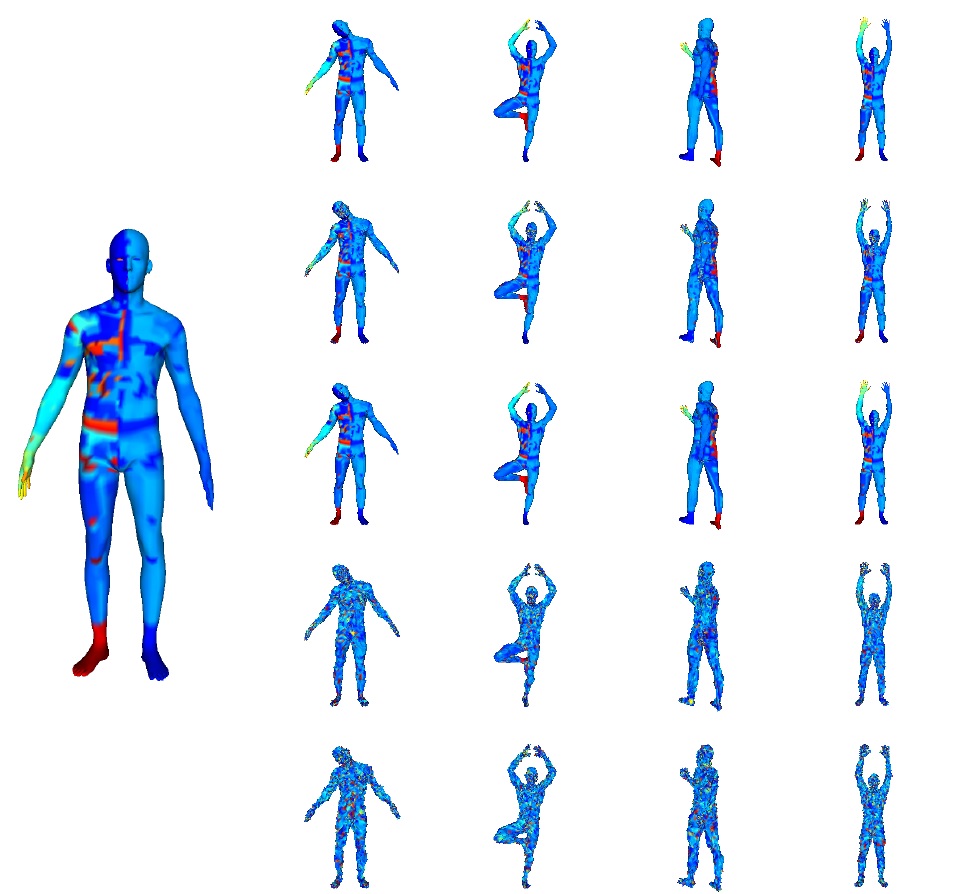}\caption{The correspondence results of testing the pre-trained model of \textit{SplineCNN}~\cite{fey2017splinecnn} on \textit{FAUST} data after adding 4 different levels of synthetic noise. The first row is \textit{FAUST} data without adding any noise. From the second row till the fifth row, the noise is added gradually from the lowest to the highest.}
								\label{fig:NoiseyFaust}
								\end {figure*}

								The registered scans of the \textit{FAUST} dataset are smooth, clean and noise-free. All the experiments reported in~\cite{fey2017splinecnn} are on the clean \textit{FAUST} data. These experiments suggest that it is important to investigate how robust to noise a given model is. \\
								In another experiment, we test on four meshes from the \textit{FAUST} dataset after adding synthetic noise of different levels 
								as depicted
								in Fig.~\ref{fig:NoiseyFaust}. The 
								initial
								row of meshes shows four of the original test meshes of \textit{FAUST} dataset without adding any noise. The following rows show different levels of added noise from level 1 (lowest noise) to level 4 (highest). The mesh shown on the left represents the reference mesh and the set of meshes on the right show the correspondence results. For level 1 noise (second row), the geometry of meshes has barely changed with respect to the original meshes (first row), except for some noise on the faces. However, the correspondence results show that there are some erroneous areas in the arms for the `hands up pose' (last pose from the left), which should not be the case since the arms did not change. Moreover, the more noisy the mesh is, the more the correspondence results get erroneous in terms of the wrongly labelled vertices and this is clear in terms if the difference in the color map between the reference mesh and the test meshes.

								\begin{figure*}[b!]
									\centering
									\includegraphics[width=0.8\columnwidth]{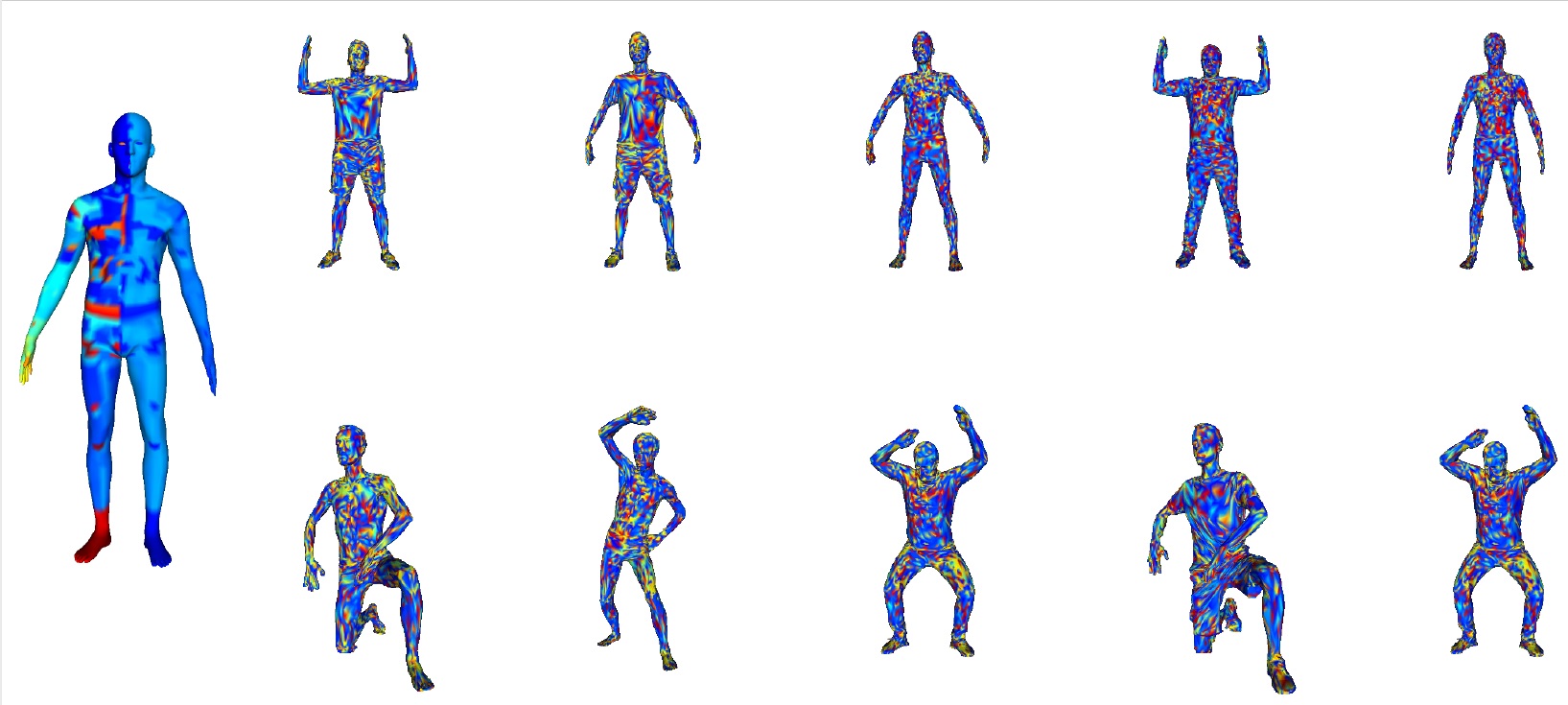}\caption{Testing the pre-trained model of \textit{SplineCNN}~\cite{fey2017splinecnn} on some clothed and unclothed scans from the \textit{3DBodyTexdata}~\cite{saint20183dbodytex} dataset.}
									\label{fig:Shapify}
									\end {figure*}

									Real mesh data is more challenging because it is more noisy, incomplete and less regular with variable sampling. Because of the noise, the vertices of an input mesh do not have an exact correspondence with the template mesh, making the correspondence ambiguous. Also, the connectivity pattern usually varies from mesh to another even if re-sampled to the same number of vertices. Possible holes in the data make it hard to find the exact correspondence to the reference mesh. Variability in the sampling requires robust methods that can adapt to different scales and handle different levels of features. Recently, \textit{Monte Carlo Convolution}~\cite{hermosilla2018monte} is proposed to handle sampling at different levels for point clouds learning. The authors in this work propose to represent the convolution kernel as a ``\textit{Multi-Layer Perceptron} (MLP)" where the convolution is formulated as a Monte Carlo intergration problem. This notion enables to combine information of the point cloud from multiple samplings at different levels, where Poisson disk sampling is used as a scalable means of hierarchical point cloud learning. This showed robustness even when all the training data is non-uniformly samples. This method achieves relatively better results compared to \textit{PointNet++}~\cite{qi2017pointnet++}. 
									
									To test for the robustness of the \textit{SplineCNN} model with respect to all of these challenges, we test on real-world clothed and unclothed scans from the \textit{3DBodyTex} dataset~\cite{saint20183dbodytex} as shown in Fig.~\ref{fig:Shapify}. We down-sample 10 meshes of the \textit{3DBodyTex} dataset to 6890 vertices to be equivalent to the number of vertices of \textit{FAUST} dataset. However, the down-sampled data have different connectivity pattern from the \textit{FAUST} data, which makes the task of the correspondence harder. As depicted in Fig.~\ref{fig:Shapify}, \textit{SplineCNN} model is not able to handle the \textit{3DBodyTex} data which is of different topology. This results in highly erroneous correspondence results where the resulting color map on the test data is very far from the color map of the reference mesh.

									\section{Conclusion}\label{conclusion}
									
									The ongoing evolution of scanning devices caused a huge increase in the amount of 3D data available in the 3D computer vision research community. This opens the door for new opportunities investigating the properties of different 3D objects and learning their geometric properties despite challenges imposed by the data itself. Fortunately, DL techniques revolutionized the learning performance on various 2D computer vision tasks which encouraged the 3D research community to adopt the same path. However, extending 2D DL to 3D data is not a straightforward tasks depending on the data representation itself and the task at hand. In this work, we categorized the 3D data representations based on their internal structure to Euclidean and non-Euclidean representations. Following the same classification, we discussed different DL techniques applied to 3D data based on the data representation and how the internal structure of the data is treated. In the Euclidean DL family, the reported results in the literature shows that multi-view representations achieve the state-of-the-art classification performance and outperforms other methods that exploit the full geometry of the 3D shape (i.e., volumetric methods) providing a more efficient way to learn the properties of 3D shapes. On the other branch of the non-Euclidean DL techniques, results are reported near perfect on the correspondence task in various recent papers such as~\cite{fey2017splinecnn,verma2018feastnet}. These correspondence experiments were carried on clean, smooth and ideal data. 
									In this paper, state-of-the-art \textit{SplineCNN}~\cite{fey2017splinecnn} method have been tested over different dataset, under different different conditions that emulate the real-world scenarios.
									The obtained results showed that, even with the same topology and similar poses, this model does not generalize to new or noisy data.  There is clearly a need to further investigate ways to improve the robustness of 3D DL models and ensure their generalization to real data while exploiting the different existent representations of 3D data.

									
									\section{Acknowledgement}                                                  
									This work has been funded by FNR project IDform under the agreement CPPP17/IS/11643091/IDform/\\Aouada, Luxembourg and by Artec Europe SARL.

									\bibliographystyle{ACM-Reference-Format}
									\bibliography{BibJournal}

\end{document}